\documentclass[11pt,a4paper]{article}
\usepackage[hyperref]{emnlp2020}
\usepackage{times}
\usepackage{latexsym}

\usepackage{microtype}
\usepackage{amsmath}
\usepackage{bm}
\usepackage{color}
\usepackage{multirow}
\usepackage{amsmath}
\usepackage{subfig}
\usepackage{amsfonts}
\usepackage{multirow}
\usepackage{makecell}
\usepackage{arydshln}
\usepackage{graphicx} 
\usepackage{graphics}
\usepackage{bbm}
\usepackage{balance}
\usepackage{algorithm}
\usepackage[noend]{algpseudocode}
\usepackage{booktabs}
\usepackage{textcomp}
\usepackage{threeparttable}
\usepackage{paralist}
\usepackage[normalem]{ulem}

\usepackage{CJKutf8}
\usepackage{booktabs}

\usepackage{cleveref}
\crefformat{section}{\S#2#1#3} 

\aclfinalcopy 
\def\aclpaperid{2674} 

\definecolor{zptu}{RGB}{18, 141, 21}

\definecolor{jiao}{RGB}{168, 0, 128}

\title{Data Rejuvenation: Exploiting Inactive Training Examples for \\Neural Machine Translation}

\author{
Wenxiang Jiao$^\dagger$\thanks{~~Work was mainly done when Wenxiang Jiao and Shilin He were interning at Tencent AI Lab.}  ~~  Xing Wang$^\ddagger$  ~~ Shilin He$^\dagger$ ~~ Irwin King$^\dagger$  ~~ Michael R. Lyu$^\dagger$ ~~Zhaopeng Tu$^\ddagger$ \\
{$^\dagger$Department of Computer Science and Engineering} \\
{The Chinese University of Hong Kong, HKSAR, China} \\
{$^\ddagger$Tencent AI Lab} \\
$^\dagger${\tt \{wxjiao,slhe,king,lyu\}@cse.cuhk.edu.hk} \\
$^\ddagger${\tt \{brightxwang,zptu\}@tencent.com} \\}

\date{}

\begin{document}
\maketitle
\begin{abstract}
Large-scale training datasets lie at the core of the recent success of neural machine translation (NMT) models. However, the complex patterns and potential noises in the large-scale data make training NMT models difficult. In this work, we explore to identify the inactive training examples which contribute less to the model performance, and show that the existence of inactive examples depends on the data distribution. We further introduce \emph{data rejuvenation} to improve the training of NMT models on large-scale datasets by exploiting inactive examples. The proposed framework consists of three phases. 
First, we train an {\em identification model} on the original training data, and use it to distinguish inactive examples and active examples by their sentence-level output probabilities.
Then, we train a {\em rejuvenation model} on the active examples, which is used to re-label the inactive examples with forward-translation. Finally, the rejuvenated examples and the active examples are combined to train the final NMT model. Experimental results on WMT14 English-German and English-French datasets show that the proposed \emph{data rejuvenation} consistently and significantly improves performance for several strong NMT models. Extensive analyses reveal that our approach stabilizes and accelerates the training process of NMT models, resulting in final models with better generalization capability.~\footnote{The source code is available at \url{https://github.com/wxjiao/Data-Rejuvenation}}

\end{abstract}

\section{Introduction}

Neural machine translation (NMT) is a data-hungry approach, which requires a large amount of data to train a well-performing NMT model~\cite{koehn2017six}. However, the complex patterns and potential noises in the large-scale data make training NMT models difficult. 
To relieve this problem, several approaches have been proposed to better exploit the training data, such as curriculum learning~\cite{platanios:2019:NAACL}, data diversification~\cite{nguyen2019data}, and data denoising~\cite{wang2018denoising}.

In this paper, we explore an interesting alternative which is to reactivate the {\em inactive examples} in the  training data for NMT models. By definition, inactive examples are the training examples that only marginally contribute to or even inversely harm the performance of NMT models. 
Concretely, we use sentence-level output probability~\cite{kumar2019calibration} assigned by a trained NMT model to measure the activeness level of training examples, and regard the examples with the least probabilities as inactive examples (\cref{sec:identification}). Experimental results show that removing 10\% most inactive examples can marginally improve translation performance. In addition, we observe a high overlapping ratio (e.g., around 80\%) of the most inactive and active examples across random seeds, model capacity, and model architectures (\cref{sec:exp_identification}).
These results provide empirical support for our hypothesis of the existence of inactive examples in large-scale datasets, which is invariant to specific NMT models and depends on the data distribution itself.

We further propose \emph{data rejuvenation} to rejuvenate the inactive examples to improve the performance of NMT models. 
Specifically, we train an NMT model on the active examples as the rejuvenation model to re-label the inactive examples, resulting in the rejuvenated examples~(\cref{sec:rejuvenation}). The final NMT model is trained on the combination of the active examples and rejuvenated examples.
Experimental results show that the data rejuvenation approach consistently and significantly improves performance on SOTA NMT models (e.g., \textsc{Lstm}~\cite{domhan2018much}, \textsc{Transformer}~\cite{Vaswani:2017:NIPS}, and \textsc{DynamicConv}~\cite{wu2019pay}) on the benchmark WMT14 English-German and English-French datasets~(\cref{sec:main}).
Encouragingly, our approach is also complementary to existing data manipulation methods (e.g., data diversification~\cite{nguyen2019data} and data denoising~\cite{wang2018denoising}), and combining them can further improve performance.

\begin{figure*}[t]
    \centering
    \includegraphics[width=0.75\textwidth]{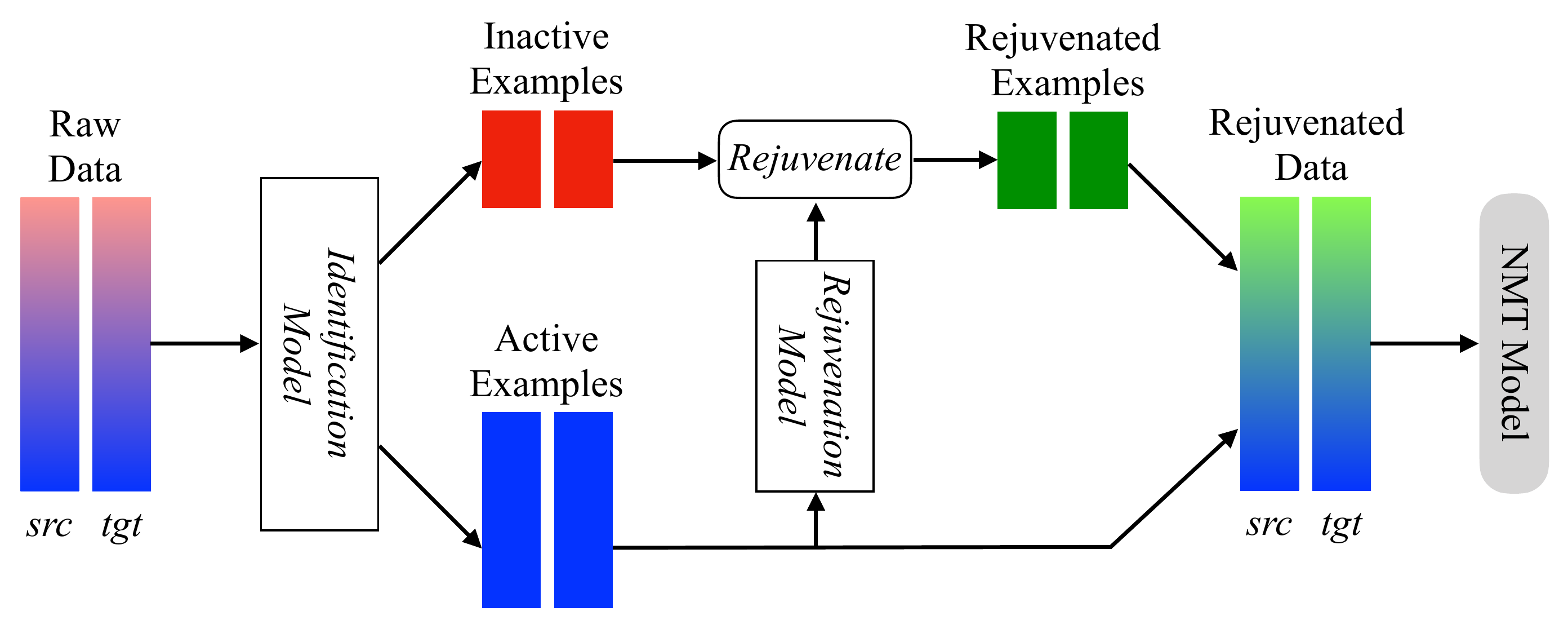}
    \caption{The framework of data rejuvenation.
    The inactive examples from the original training data are identified by the {\em identification model}, then rejuvenated by the {\em rejuvenation model}. The rejuvenated examples along with the active examples are used together to train the NMT model. Best view in color.}
    \label{fig:framework}
\end{figure*}

Finally, we conduct extensive analyses to better understand the inactive examples and the proposed data rejuvenation approach. Quantitative analyses reveal that the inactive examples are more difficult to learn than active ones, and rejuvenation can reduce the learning difficulty~(\cref{sec:linguistic}). The rejuvenated examples stabilize and accelerate the training process of NMT models~(\cref{sec:learning}), resulting in final models with better generalization capability~(\cref{sec:generalization}).

Our contributions of this work are as follows: 
\begin{itemize}
\item Our study demonstrates the existence of inactive examples in large-scale translation datasets, which mainly depends on the data distribution.
\item We propose a general framework to rejuvenate the inactive examples to improve the training of NMT models.
\end{itemize}

\section{Related Work}
\label{sec:related_work}

\paragraph{Data Manipulation.}
Our work is closely related to previous studies on manipulating training data for NMT models, which focuses on exploiting the original training data without augmenting additional data. For example, the data denoising approach~\cite{wang2018denoising} aims to identify and clean the noise training examples. Data diversification~\cite{nguyen2019data} tries to diversify the training data by applying forward-translation~\cite{Zhang:2016:EMNLP} to the source side of the parallel data, or back-translation~\cite{Sennrich:BT} to the target side of parallel data in a reverse translation direction.
Our approach is complementary to theirs, and using them together can further improve translation performance (Table~\ref{tab:comparison}). 
Another distantly related direction is to simplify the source sentences so that a black-box machine translation system can better translate them~\cite{mehta:2020:AAAI}, which is out of scope in this work.

\paragraph{Distinguishing Training Examples.}
Our work is also related to previous work on distinguishing training examples in machine learning. 
One stream is to re-weight training examples with different choices of  preferred examples during the training stage. For example, self-paced learning~\cite{kumar2010self} prefers easy examples, hard example mining~\cite{shrivastava2016training} exploits hard examples, and active learning~\cite{chang2017active} emphasizes high variance examples.
Another stream is to schedule the order of training examples according to their difficulty, e.g., curriculum learning which has been applied to the training of NMT models successfully~\cite{kocmi:2017:RANLP, zhang:2018:arxiv, platanios:2019:NAACL,Wang:2019:ACL2,liu:2020:ACL}. 
In contrast, we explore strategies to simplify the difficult (i.e., inactive) examples without changing the model architecture and model training strategy.

\paragraph{Inactive Examples in Computer Vision Dataset.} \newcite{birodkar2019semantic} reveals that data redundancy exists in large-scale image recognition datasets, e.g., CIFAR-10~\cite{krizhevsky2009learning} and ImageNet~\cite{deng2009imagenet} datasets. They find that a subset can generalize on par with the full dataset and that at least 10$\%$ of training data are redundant in these large-scale image classification datasets. 
Our results confirm these findings on the large-scale NLP datasets. In addition, we propose to rejuvenate the inactive examples to further improve the model performance.

\section{Methodology}

Figure~\ref{fig:framework} shows the framework of the \emph{data rejuvenation} approach, in which we introduce two models: an identification model and a rejuvenation model. The {\em identification model} distinguishes the inactive examples from the active ones. The {\em rejuvenation model}, which is trained on the active examples, rejuvenates the inactive examples. The rejuvenated examples and the active examples are combined to train the final NMT model.

There are many possible ways to implement the general idea of data rejuvenation. The aim of this paper is not to explore this whole space but simply to show that one fairly straightforward implementation works well and that data rejuvenation helps.

\subsection{Identification Model}
\label{sec:identification}

We describe a simple heuristic to implement the identification model by leveraging the output probabilities of NMT models.
The training objective of the NMT model is  to maximize the log-likelihood of the training data$\{\left[{\bf x}^n, {\bf y}^n\right]\}_{n=1}^{N}$:
\begin{eqnarray}
L(\theta)        =  \sum_{n=1}^{N} \log P(\mathbf{y}^{n} | \mathbf{x}^{n}) .
\end{eqnarray}
The trained NMT model assigns a sentence-level probability $P(\mathbf{y}|\mathbf{x})$ to each sentence pair $(\mathbf{x}, \mathbf{y})$, indicating the confidence of the model to generate the target sentence $\mathbf{y}$ from the source one $\mathbf{x}$~\cite{kumar2019calibration,Wang:2020:ACL}. Intuitively, if a training example has a low sentence-level probability, it is less likely to provide useful information for improving model performance, and thus is regarded as an inactive example. 

Therefore, we adopt sentence-level probability $P(\mathbf{y}|\mathbf{x})$ as the metric to measure the activeness level of each training example:
\begin{equation}
\label{eq_inactive_metric}
    I(\mathbf{y}|\mathbf{x}) =  \prod_{t=1}^T p(y_t|\mathbf{x}, \mathbf{y}_{<t}) ,
\end{equation}
where $T$ is the number of target words in the training example. $I(\mathbf{y}|\mathbf{x})$ is normalized by the length of target sentence $\mathbf{y}$ to avoid length bias.
We train an NMT model on the original training data and use it to score each training example. We treat a certain percent of training examples with the least sentence-level probabilities as inactive examples.

\subsection{Rejuvenation Model}
\label{sec:rejuvenation}

Inspired by recent successes on data augmentation for NMT, we adopt the widely-used back-translation~\cite{Sennrich:BT} and forward-translation~\cite{Zhang:2016:EMNLP} approaches to implement the rejuvenation model.
After the active examples are distinguished from the training data, we use them to train an NMT model in forward direction for forward-translation or/and reverse direction for back-translation. The trained model rejuvenates each inactive example by producing a synthetic-parallel example based on their source (for forward-translation) or target (for back-translation) side.
Benefiting from the knowledge distillation based on active examples, the rejuvenated examples consist of simpler patterns than the original examples~\cite{edunov2019evaluation}, thus are more likely to be learned by NMT models. 

\section{Experiment}

\subsection{Experimental Setup}
\label{sec:setup}

\paragraph{Data.} We conducted experiments on the  widely used WMT14 English$\Rightarrow$German 
(En$\Rightarrow$De) and English$\Rightarrow$French 
(En$\Rightarrow$Fr) datasets, which consist of about $4.5$M and $35.5$M sentence pairs, respectively. We applied BPE~\cite{Sennrich:BPE} with 32K merge operations for both language pairs. The experimental results were reported in case-sensitive BLEU score~\cite{papineni2002bleu}. 

\paragraph{Model.} 
We validated our approach on a couple of representative NMT architectures:
\begin{itemize}
    \item \textsc{Lstm}~\cite{domhan2018much} that is implemented in the \textsc{Transformer} framework.
    \item \textsc{Transformer}~\cite{Vaswani:2017:NIPS} that is based solely on attention mechanisms.
    \item \textsc{DynamicConv}~\cite{wu2019pay} that is implemented with lightweight and dynamic convolutions, which can perform competitively to the best reported \textsc{Transformer} results.
\end{itemize}
We adopted the open-source toolkit Fairseq~\cite{ott:2019:naacl} to implement the above NMT models. 
We followed the settings in the original works to train the models.
In brief, we trained the \textsc{Lstm} model for 100K steps with 32K ($4096\times8$) tokens per batch.
For \textsc{Transformer}, we trained 100K and 300K steps with 32K tokens per batch for the \textsc{Base} and \textsc{Big} models respectively.
We trained the \textsc{DynamicConv} model for 30K steps with 459K ($3584\times128$) tokens per batch.
We selected the model with the best perplexity on the validation set as the final model.

We first conducted ablation studies on the identification model (\S\ref{sec:exp_identification}) and rejuvenation model (\S\ref{sec:exp_rejuvenation}) on the WMT14 En$\Rightarrow$De dataset with \textsc{Transformer-Base}. Then we reported the translation performance on different model architectures and language pairs, as well as the comparison with previous studies (\S\ref{sec:main}).

\subsection{Identification of Inactive Examples}
\label{sec:exp_identification}

In this section, we investigated the reasonableness and consistency of the identified inactive examples.

\begin{figure}[t]
    \centering
    \subfloat[En$\Rightarrow$De]{   \includegraphics[height=0.3\textwidth]{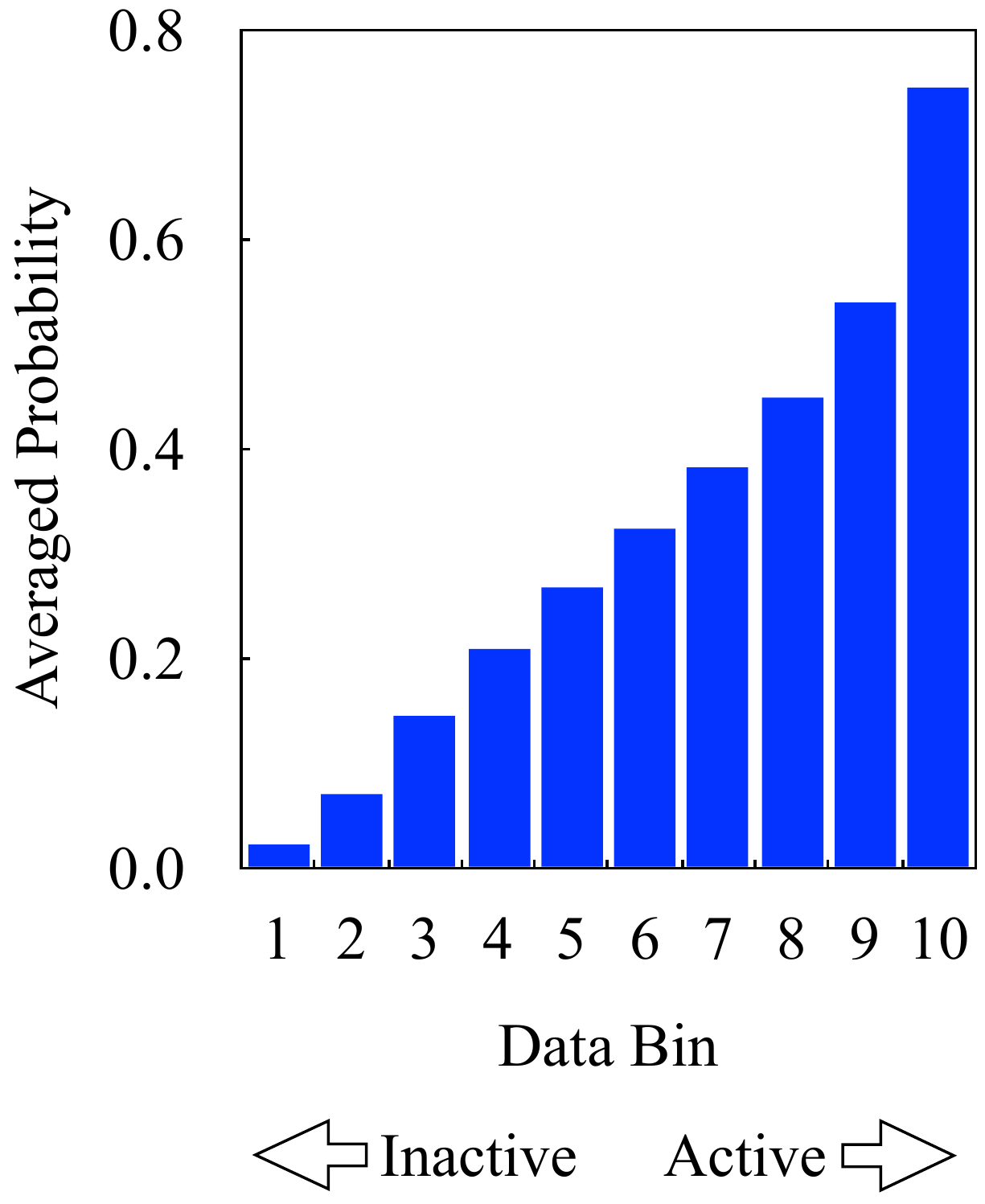}}
    \hspace{0.02\textwidth}
    \subfloat[En$\Rightarrow$Fr]{   \includegraphics[height=0.3\textwidth]{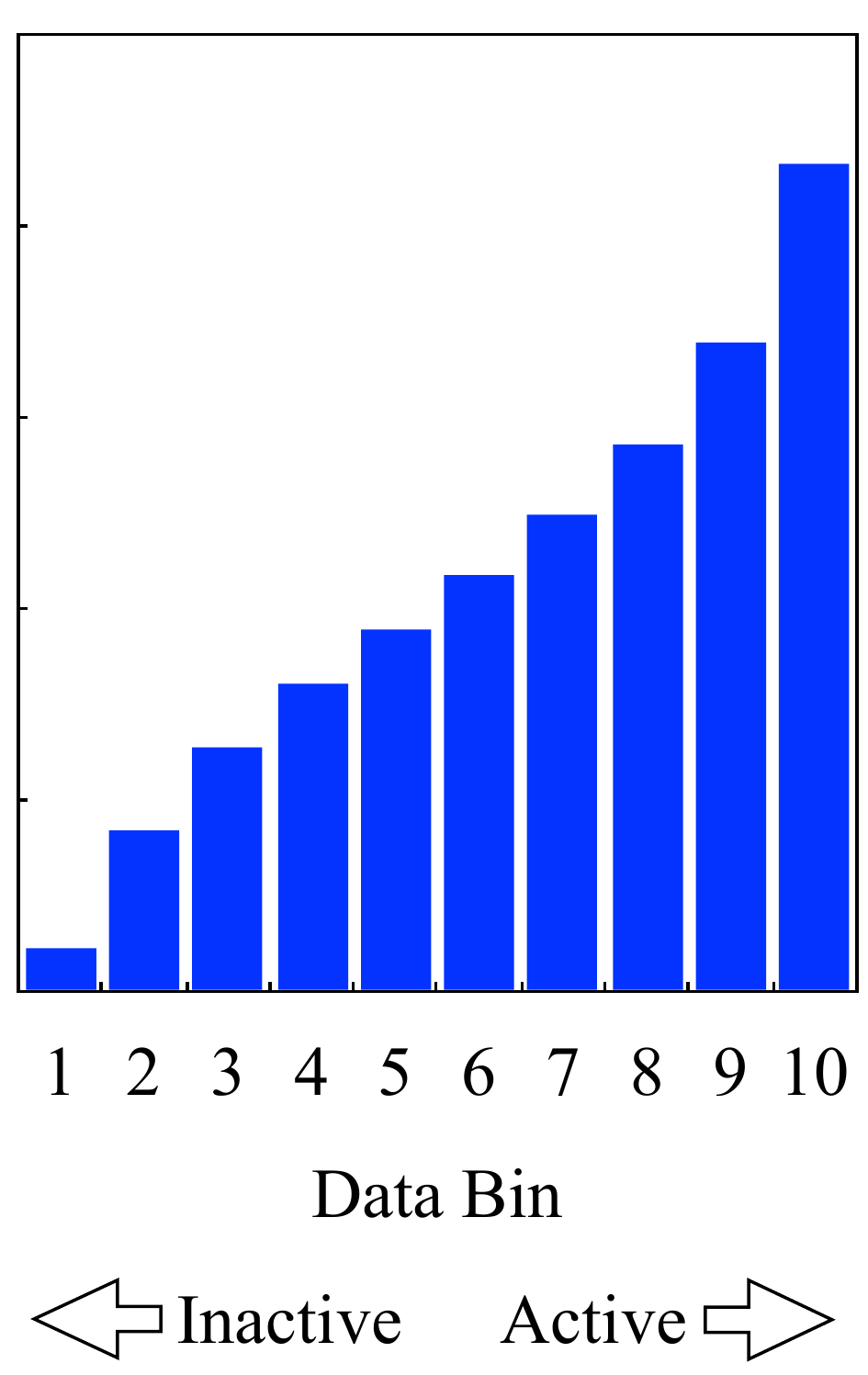}}
    \caption{Probability diagram on (a) En$\Rightarrow$De and (b) En$\Rightarrow$Fr datasets. Training examples in smaller bins (e.g., 1, 2) are regarded as inactive examples due to their lower probabilities.}
    \label{fig:probability-diagram}
\end{figure}

\paragraph{Identified Inactive Examples.}
As aforementioned, we ranked the training examples according to the sentence-level output probability (i.e., confidence) assigned by a trained NMT model. 
We followed~\newcite{Wang:2020:ACL} to partition the training examples into 10 equal bins (i.e., each bin contains 10\% of training examples) according to the ranking of their probabilities and reported the averaged probability of each bin, as depicted in Figure~\ref{fig:probability-diagram}. As seen, the examples in the $1^{st}$ data bin have much lower probabilities than the other ones, which we regard as inactive examples.

\begin{figure}[t]
    \centering
    \includegraphics[width=0.4 \textwidth]{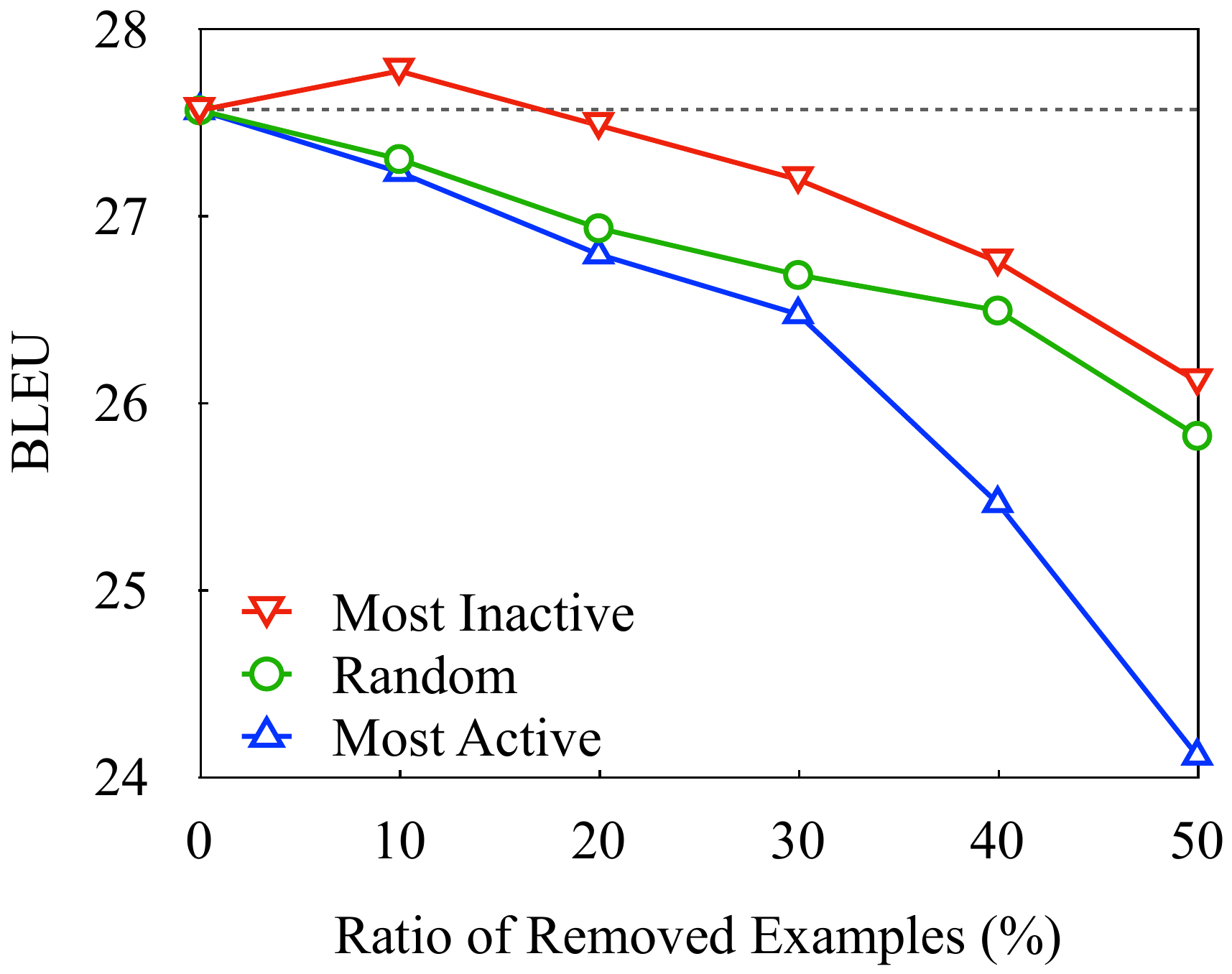}
    \caption{
    Translation performance of the NMT model trained on the training data with the most inactive examples removed.
    For comparison, results of the most active examples and randomly sampled examples are also presented.}
    \label{fig:ratio-removed-examples}
\end{figure}

\paragraph{Reasonableness of Identified Inactive Examples.}
In this experiment, we evaluated the reasonableness of the identified inactive examples by measuring their contribution to the translation performance. Intuitively, a reasonable set of inactive examples can be removed from the training data without harming the translation performance, since they cannot provide useful information to the NMT models.
Starting from this intuition, we removed a certain percentage of examples with the least probabilities (e.g., most inactive examples) from the training data, and evaluated the performance of the NMT model that is trained on the remaining data.

Figure~\ref{fig:ratio-removed-examples} shows the contribution of the most inactive examples to translation performance.  
Generally, the performance drop grows up with the increased portion of examples being removed from the training data. The declining trend of the inactive examples is more gentle than the randomly selected examples, and that of the active examples is steepest. These results demonstrate the reasonableness of the identified examples.
Encouragingly, the translation performance does not degrade when removing 10\% of the most inactive examples, which is consistent with the finding of~\newcite{birodkar2019semantic} on the CV datasets.

\begin{figure}[t]
    \centering
    \subfloat[En-De: Seed]{
    \includegraphics[height=0.30\textwidth]{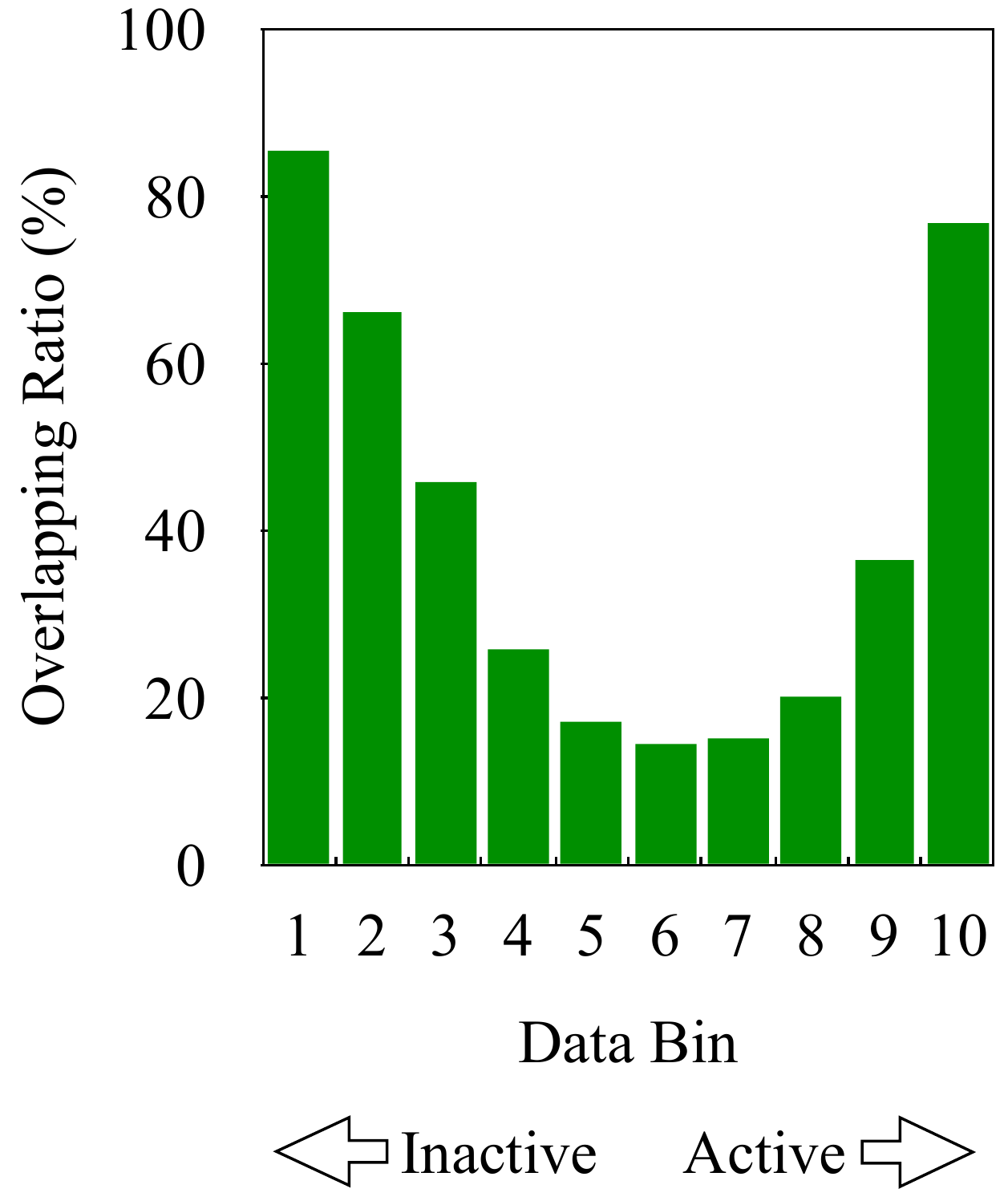}}
    \hfill
    \subfloat[En-De: Capacity]{
    \includegraphics[height=0.30\textwidth]{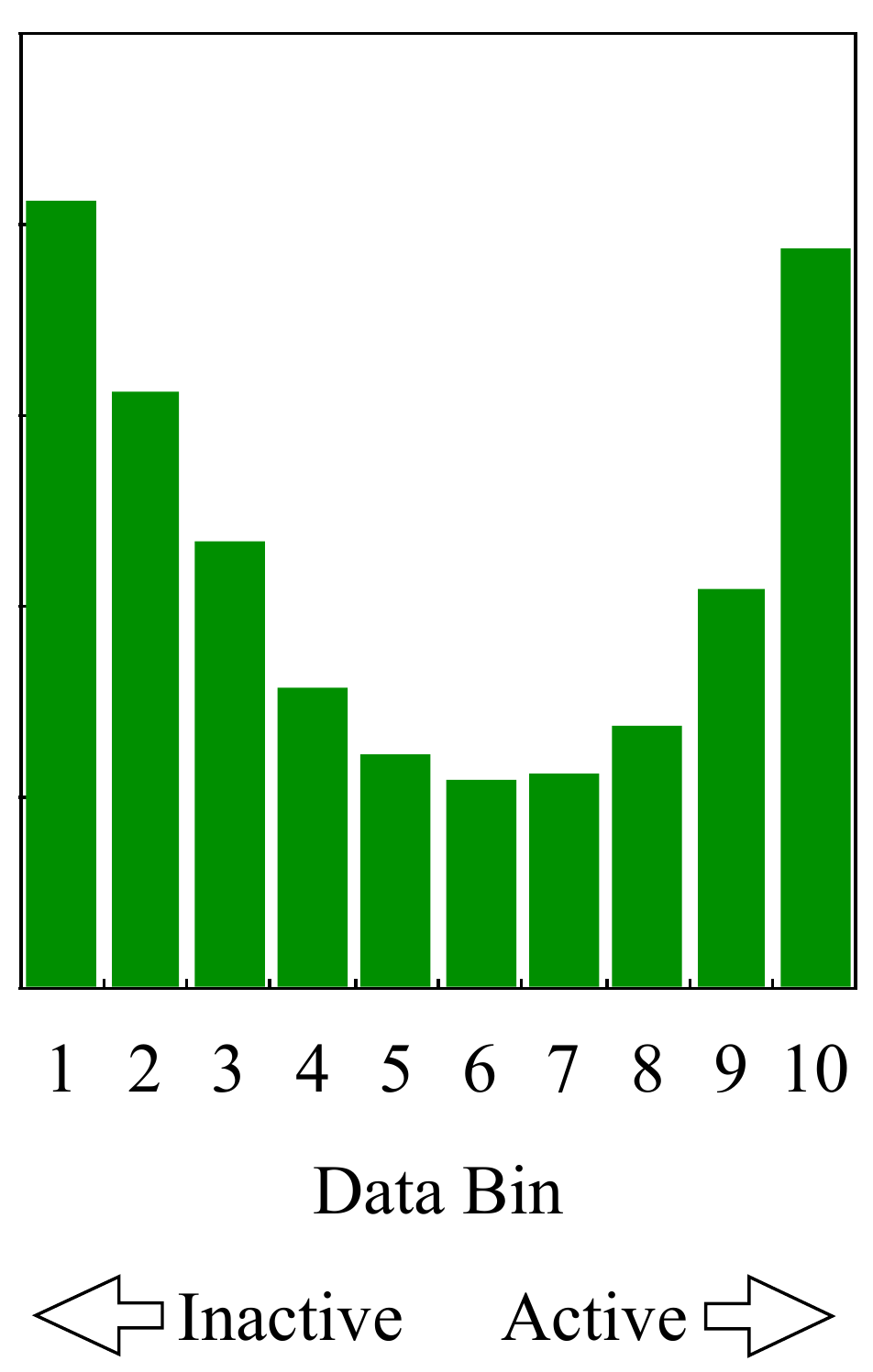}}
    \\
    \subfloat[En-De: Architecture]{
    \includegraphics[height=0.30\textwidth]{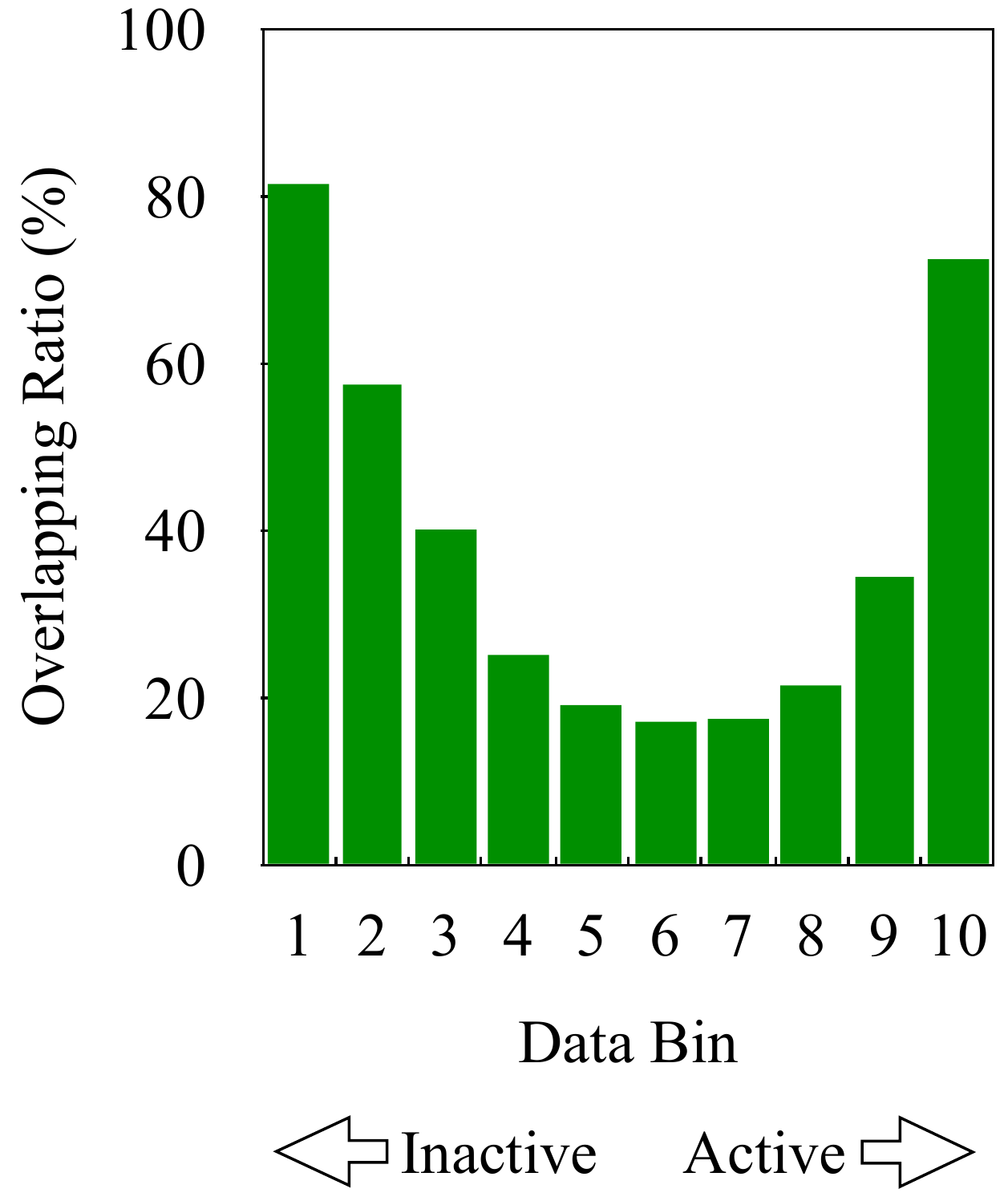}}
    \hfill
    \subfloat[En-Fr: Architecture]{
    \includegraphics[height=0.30\textwidth]{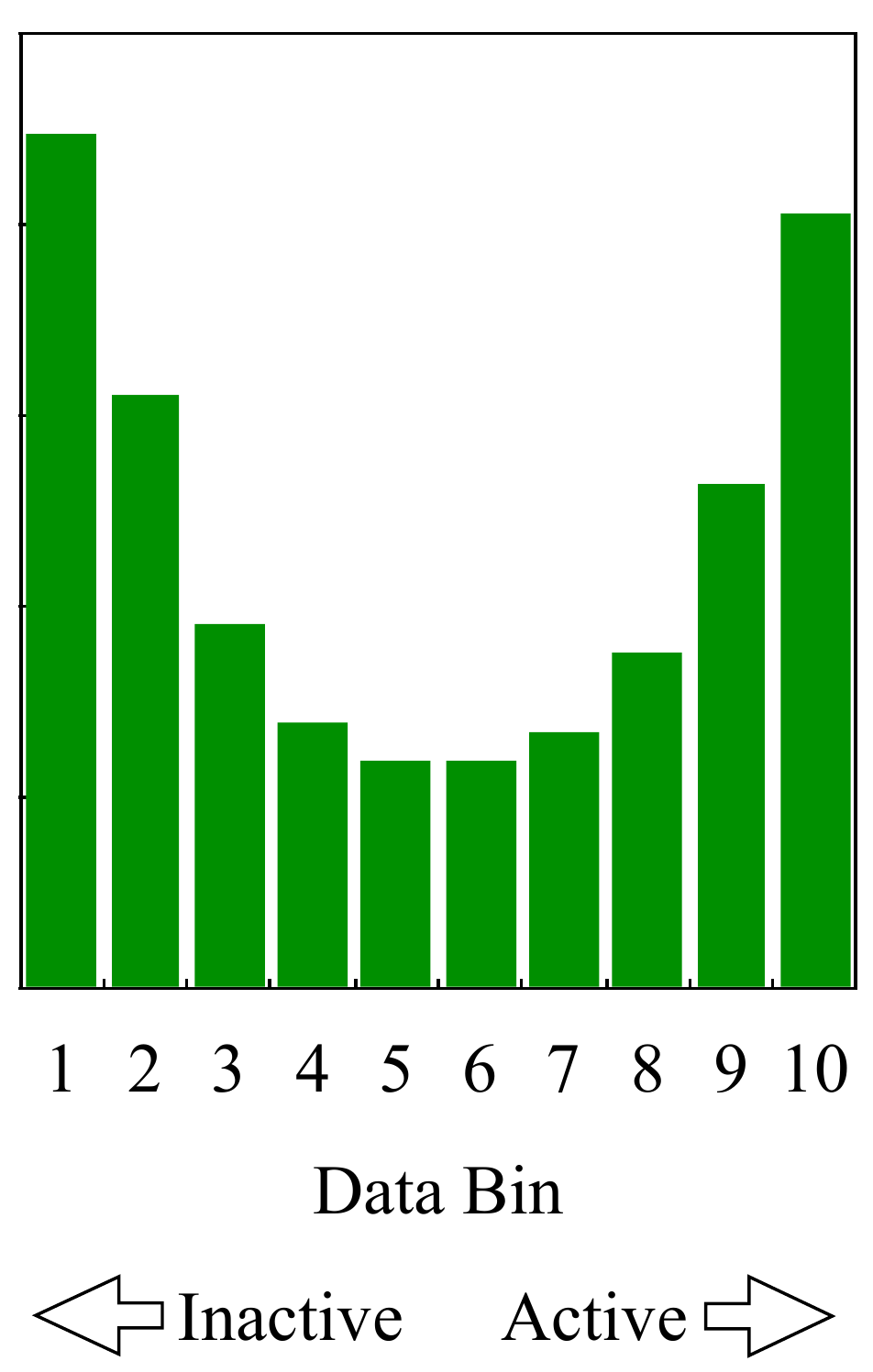}}
    \caption{Ratio of examples that are shared by different model variants: random seed (a), model capacity (b), model architecture on En$\Rightarrow$De (c) and En$\Rightarrow$Fr (d) datasets. A high overlapping ratio for most inactive examples (i.e., $1^{st}$ data bin) demonstrates that the identified inactive examples are not model-specific.}
    \label{fig:overlap}
\end{figure}

\paragraph{Consistency of Identified Inactive Examples.}
Since our identification of inactive examples relies on a pre-trained NMT model, one doubt naturally arises: {\em are the identified inactive examples model-specific}? For example, different NMT models treat different portions of the training data as the inactive examples. 
To dispel the doubt, we identified some factors that can significantly affect the performance of NMT models: 1) {\em random seeds} for \textsc{Transformer-Base}: ``1'', ``12'', ``123'', ``1234'', and ``12345''; 2) {\em model capacity} for \textsc{Transformer}: \textsc{Tiny} ($3\times256$), \textsc{Base} ($6\times512$), and \textsc{Big} ($6\times1024$); and 3) {\em model architectures}: the aforementioned architectures in Section~\ref{sec:setup}.
For each data bin, we calculated the ratio of examples that are shared by different model variants (e.g., different random seeds). Generally, a high overlapping ratio denotes the identified examples are more agreed by different models, which suggests the examples are not model-specific.

Figure~\ref{fig:overlap} depicts the results. As expected, there is always a high overlapping ratio (over 80\%) for the most inactive examples (i.e., $1^{st}$ data bin) across model variants and language pairs.
The high consistency of identified inactive examples demonstrates that {\em the proposed identification is invariant to specific models, and depends on the data distribution itself}.
Another interesting finding is that the most active examples (i.e., $10^{th}$ data bin) also holds a high agreement by model variants. 
The overlapping ratios of all model variants (i.e., seeds, capacities, and architectures, 9 models in total) on the En$\Rightarrow$De dataset are 70.9\%, and 62.5\% for the most inactive and (most) active examples, respectively. This indicates that deep learning methods share a common ability to learn from the training examples.

\begin{figure}[t]
    \centering
    \includegraphics[width=0.4\textwidth]{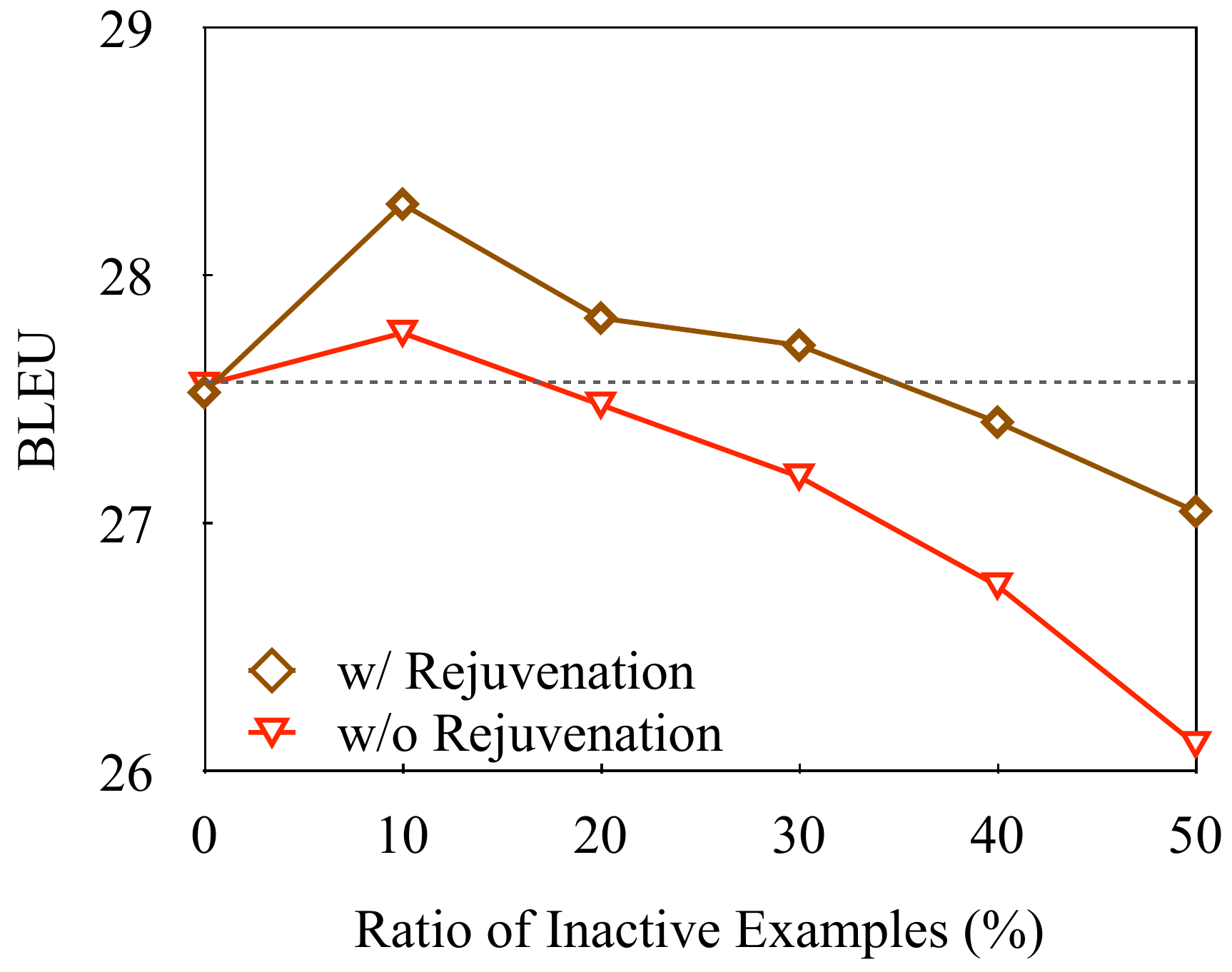}
    \caption{Effect of the ratio of examples labelled as inactive examples. We used forward-translation as the rejuvenation strategy and trained the final NMT model on the combination of rejuvenated examples and active examples from scratch.}
    \label{fig:ratio-inactive-examples}
\end{figure}

\begin{table}[t]
\centering
\begin{tabular}{c|c|c}
     \bf Rejuvenation & \bf BLEU & $\bigtriangleup$ \\
     \hline \hline
     n/a   &  27.5 & --\\
     \hline
     Forward Translation  & \bf 28.3 & +0.8\\
     Back-Translation     & 27.5 & +0.0\\
     Both                 & 27.8 & +0.3\\
  \end{tabular}
  \caption{Effect of different rejuvenation strategies.}
  \label{tab:rejuvenation-strategies}
\end{table}

\begin{table*}[t]
\setcounter{table}{2}
\centering
\begin{tabular}{c|l|| l c|| l c}
\multirow{2}{*}{\bf System}  &   \multirow{2}{*}{\bf Architecture}  & \multicolumn{2}{c}{\bf WMT14 En$\Rightarrow$De}  &  \multicolumn{2}{||c}{\bf WMT14 En$\Rightarrow$Fr}\\
    \cline{3-6}
        &   &   BLEU & $\bigtriangleup$    & BLEU & $\bigtriangleup$ \\
    \hline \hline
    \multicolumn{6}{c}{{\em Existing NMT Systems}} \\
    \hline
    \multirow{2}{*}{\newcite{Vaswani:2017:NIPS}} &   \textsc{Transformer-Base}    &    27.3  & --  &  38.1 &  -- \\
     &  \textsc{Transformer-Big} &    28.4  & --  &  41.0 &  -- \\
    \hdashline
    \newcite{ott2018scaling} & \textsc{Scale Transformer} &  29.3  &  --  &  43.2  &  --\\
    \hdashline
    \newcite{wu2019pay} & \textsc{DynamicConv} & 29.7 & -- &  43.2 & --\\
    \hline
    \hline
    \multicolumn{6}{c}{{\em Our NMT Systems}}   \\ 
    \hline
    \multirow{10}{*}{\em This work}
    &   \textsc{Lstm}              &    26.5   &   --  &   40.6  &   --\\
    &   ~~~ +  Data Rejuvenation   &    27.0$^\uparrow$       &    +0.5   &  41.1$^\uparrow$  &   +0.5\\ 
    \cline{2-6}
    &   \textsc{Transformer-Base}  &  27.5  & -- &  40.2  &   -- \\
    &   ~~~ +  Data Rejuvenation   &  28.3$^\Uparrow$  & +0.8  &  41.0$^\Uparrow$ & +0.8 \\ 
    \cdashline{2-6}
    &   \textsc{Transformer-Big}  &  28.4  & -- &   42.4 &   -- \\
    &   ~~~ +  Data Rejuvenation  &  29.2$^\Uparrow$  &  +0.8   & 43.0$^\uparrow$  &  +0.6 \\ 
    \cdashline{2-2} \cdashline{3-4} \cdashline{5-6}
    &   ~~~ + Large Batch  &   29.6     & -- &  43.5  &   -- \\
    &   ~~~~~~~~ +  Data Rejuvenation  &  30.3$^\Uparrow$  &  +0.7 & 44.0$^\uparrow$ & +0.5  \\
    \cline{2-6}
    &   \textsc{DynamicConv}      &   29.7             & --     & 43.3 &   -- \\
    &   ~~~ +  Data Rejuvenation  &   30.2$^\uparrow$  &  +0.5  & 43.9$^\uparrow$  &  +0.6 \\
  \end{tabular}
  \caption{Evaluation of translation performance across model architectures and language pairs. ``$\uparrow/\Uparrow$'':  indicate statistically significant improvement over the corresponding baseline $p < 0.05/0.01$ respectively.}  
  \label{tab:main}
\end{table*}

\begin{table}[t]
\setcounter{table}{1}
\fontsize{10}{11}\selectfont
\centering
\begin{tabular}{l|c|c}
     \bf Training Data  & \bf BLEU & $\bigtriangleup$ \\
     \hline \hline
     Raw Data   &  27.5 & --\\
     \hline
     - 10\% {\em Inactive} Examples  & 27.8  &  +0.3\\
     ~~~+ Rejuvenated Examples & 28.3  &  +0.8\\
     \hline
     - 10\% {\em Random} Examples  & 27.4  &  -0.1\\
     ~~~+ Rejuvenated Examples  & 27.3  &  -0.2\\
  \end{tabular}
  \caption{Comparing data rejuvenation on identified inactive examples and forward translation on randomly sampling examples.}
  \label{tab:training-strategies}
\end{table}

\subsection{Rejuvenation of Inactive Examples}
\label{sec:exp_rejuvenation}

In this section, we evaluated the impact of different components on the rejuvenation model. 

\paragraph{Ratio of Examples Labelled as Inactive.}
After all examples were assigned a sentence-level probability by the identification model, we labelled $R\%$ of examples with the least probabilities as the inactive examples. 
We investigated the effect of different $R$ on translation performance, as shown in Figure~\ref{fig:ratio-inactive-examples}. Clearly, rejuvenating the inactive examples consistently outperforms its non-rejuvenated counterpart, demonstrating the necessity of the data rejuvenation. Concerning the rejuvenation model, the BLEU score decreases with the increase of $R$. This is intuitive, since examples with relative higher probabilities (e.g., beyond the 10\% most inactive examples) can provide useful information for NMT models, and rejuvenating them would inversely harm the translation performance.
In the following experiments, we treat 10\% examples with the least probabilities as inactive examples.

\paragraph{Effect of Rejuvenation Strategy.} Table~\ref{tab:rejuvenation-strategies} lists the results of different rejuvenation strategies. Surprisingly, the back-translation strategy does not improve performance.
{One possible reason is that the inactive examples are identified by a forward-translation model (\S\ref{sec:identification}), indicating that these inactive examples are more difficult for NMT models to generate from the source side to the target side, rather than in the reverse direction.}
We conjecture that forward translation strategy may alleviate this problem by constructing a synthetic example, in which each source side is associated with a simpler target side.
Combining both strategies cannot further improve translation performance.
In the following experiments, we use forward translation as the default rejuvenation strategy.

\paragraph{Benefiting from Forward Translation or Data Rejuvenation?} Some researchers may doubt: {\em does the improvement indeed come from data rejuvenation, or just from forward translation?} 
To dispel the doubt, we conducted the comparison experiment by randomly selecting 10\% training examples as inactive examples and applying data rejuvenation with forward translation strategy.
As shown in Table~\ref{tab:training-strategies}, removing 10\% random examples inversely harms the translation performance, and rejuvenating them leads to a further decrease of performance. In contrast, the proposed data rejuvenation improves performance as expected. These results provide empirical support for our claim that the improvement comes from the proposed data rejuvenation rather than forward translation.

\subsection{Main Results}
\label{sec:main}

Table~\ref{tab:main} lists the results across model architectures and language pairs. 
Our \textsc{Transformer} models achieve better results than that reported in previous work~\cite{Vaswani:2017:NIPS}, especially on the large-scale En$\Rightarrow$Fr dataset (e.g., more than 1.0 BLEU points).
~\newcite{ott2018scaling} showed that models of larger capacity benefit from training with large batches. Analogous to \textsc{DynamicConv}, we trained another \textsc{Transformer-Big} model with 459K tokens per batch (``+ Large Batch'' in Table~\ref{tab:main}) as a strong baseline.
We tested statistical significance with paired bootstrap resampling~\cite{Koehn2004:emnlp} using \texttt{compare-mt}\footnote{\url{https://github.com/neulab/compare-mt}}~\cite{neubig2019:naacl}.

Clearly, our data rejuvenation consistently and significantly improves translation performance in all cases, demonstrating the effectiveness and universality of the proposed data rejuvenation approach.
It's worth noting that our approach achieves significant improvements without introducing any additional data and model modification. It makes the approach robustly applicable to most existing NMT systems.

\begin{table}[t]
\setcounter{table}{3}
\fontsize{10}{11}\selectfont
\centering
\begin{tabular}{l| c |  c}
    {\bf Model} & {\bf BLEU} &  $\bigtriangleup$ \\
    \hline \hline
    \textsc{Transformer-Base}        &   27.5  &  -- \\
    ~~~ + {\em Data Rejuvenation}    &   28.3  &  +0.8 \\
    \hline
    ~~~ + Data Diversification-BT       &   26.9  &  -0.6\\
    ~~~~~~~~ + {\em Data Rejuvenation}  &   27.9  &  +0.4\\
    \hline
    ~~~ + Data Diversification-FT       &   28.1  &   +0.6\\
    ~~~~~~~~ + {\em Data Rejuvenation}  &   28.5  &   +1.0\\
    \hline
    ~~~ + Data Denoising             &   28.1  &   +0.6\\
    ~~~~~~~~ + {\em Data Rejuvenation}  &   28.6  &   +1.1\\
    \end{tabular}
    \caption{Comparison with other data manipulation approaches. Results are reported on the En$\Rightarrow$De test set.} 
    \label{tab:comparison}
\end{table}

\paragraph{Comparison with Previous Work.}

The proposed \emph{data rejuvenation} approach belongs to the family of data manipulation. 
Accordingly, we compare it with several widely-used manipulation strategies: data diversification~\cite{nguyen2019data}, and data denoising~\cite{wang2018denoising}.

For data diversification, we used both forward-translation~\citep[FT,][]{Zhang:2016:EMNLP} and back-translation~\citep[BT,][]{Sennrich:BT} strategies on the original training data, and no monolingual data is introduced. The final NMT model was trained on the combination of the original and the synthetic parallel data. Our approach is similar to ``Data Diversification-FT'' except that we only forward-translate the identified inactive examples (10\% of the training data), while they forward-translate all the training examples.

For data denoising, we ranked the training data according to a noise metric, which requires a set of trusted examples. Following~\newcite{wang2018denoising}, we used WMT newstest 2010-2011 as the trusted data, which consists of 5492 examples. The trained NMT model on the raw data was regarded as the noisy model, which was then fine-tuned on the trusted data to obtain the denoised model. For each sentence pair, a noise score is computed based on the noisy and denoised models, which is used for instance sampling during training.

Table~\ref{tab:comparison} shows the comparison results on the WMT14 En$\Rightarrow$De test set.
All approaches improve translation performance individually except for data diversification with back-translation. 
Our approach can obtain further improvement on top of these manipulation approaches, indicating that data rejuvenation is complementary to them.

In addition, we computed the overlapping ratio between the noisiest and most inactive examples (10\% of the training data) identified by data denoising and data rejuvenation approaches, respectively.
We found that there are only 32\% of examples that are shared by the two approaches, indicating that the inactive examples are not necessarily noisy examples. 
In order to better understand the characteristics of inactive examples, we will give more detailed analyses on linguistic properties of the inactive examples in Section~\ref{sec:linguistic}.

\section{Analysis and Discussion}
\label{sec:analysis}

In this section, we performed an extensive study to understand inactive examples and data rejuvenation in terms of linguistic properties (\S\ref{sec:linguistic}), learning stability (\S\ref{sec:learning}) and generalization capacity (\S\ref{sec:generalization}). We also investigated the strategy to speed up the pipeline of data rejuvenation (\S\ref{sec:speedup}).
Unless otherwise stated, all experiments were conducted on the En$\Rightarrow$De dataset with \textsc{Transformer-Base}.

\subsection{Linguistics Properties}
\label{sec:linguistic}

\begin{figure}[t]
    \centering
    \subfloat[Frequency Rank]{        \includegraphics[height=0.2\textwidth]{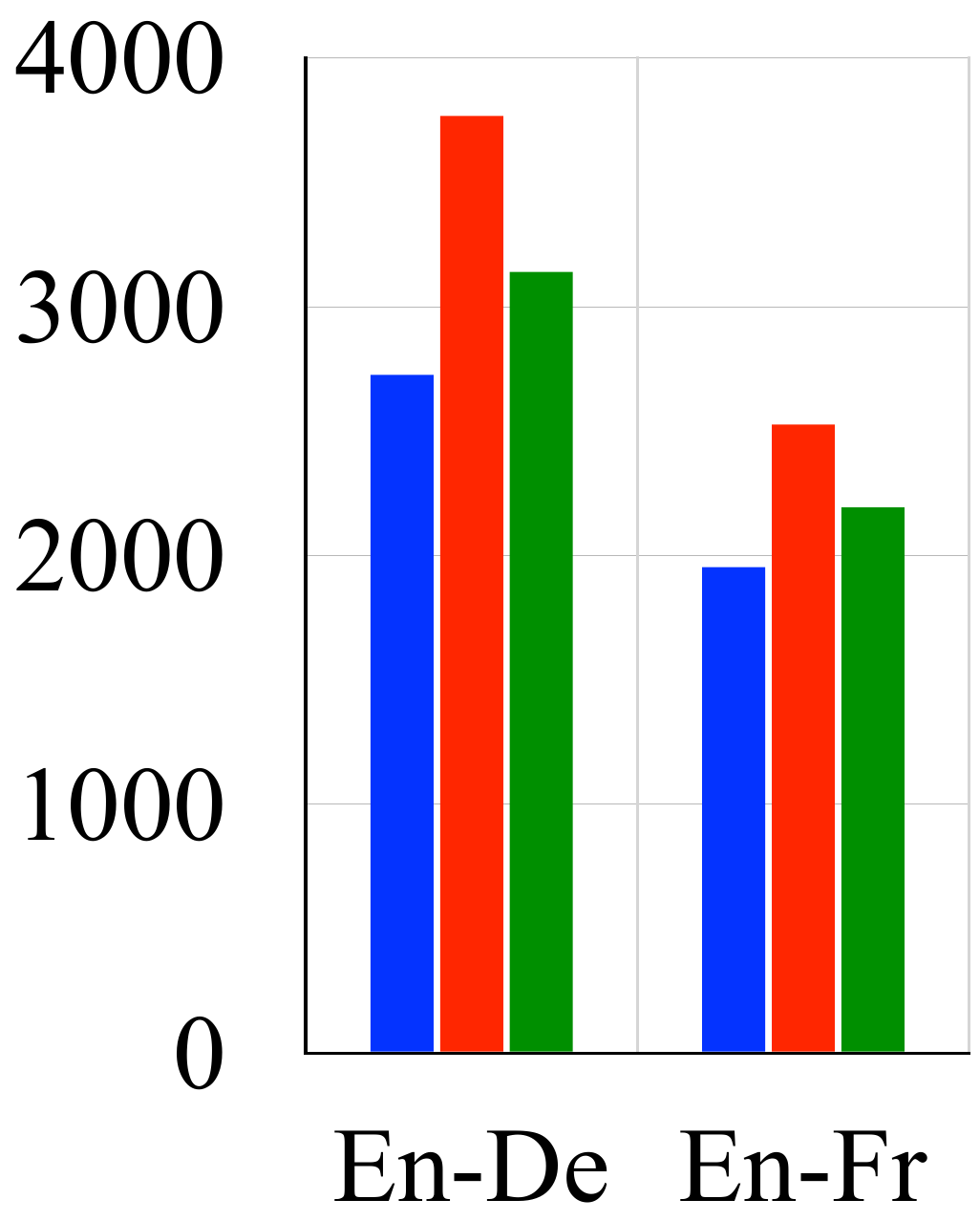}} \hfill
    \subfloat[Coverage]{        \includegraphics[height=0.2\textwidth]{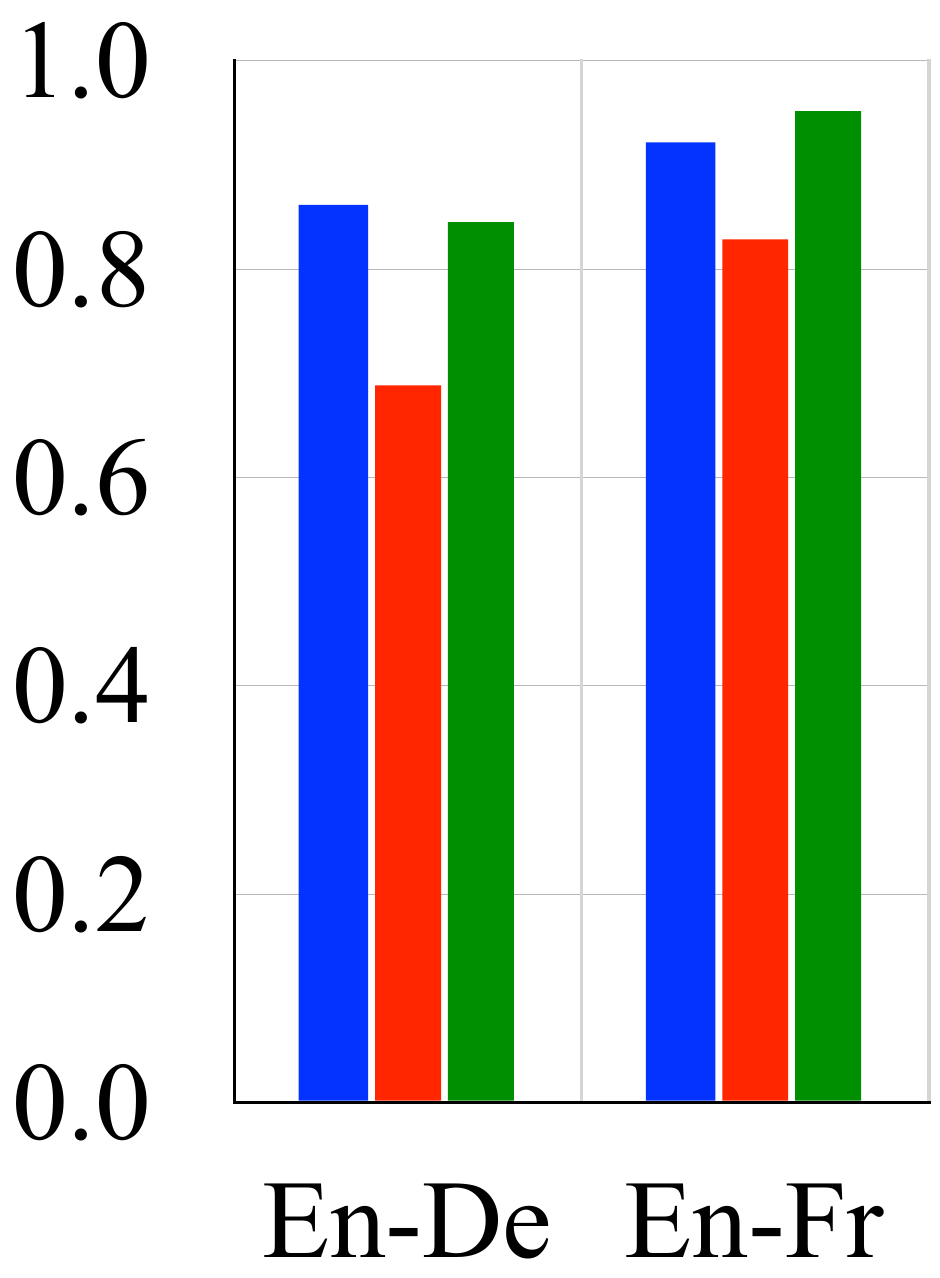}} \hfill
    \subfloat[Uncertainty]{        \includegraphics[height=0.2\textwidth]{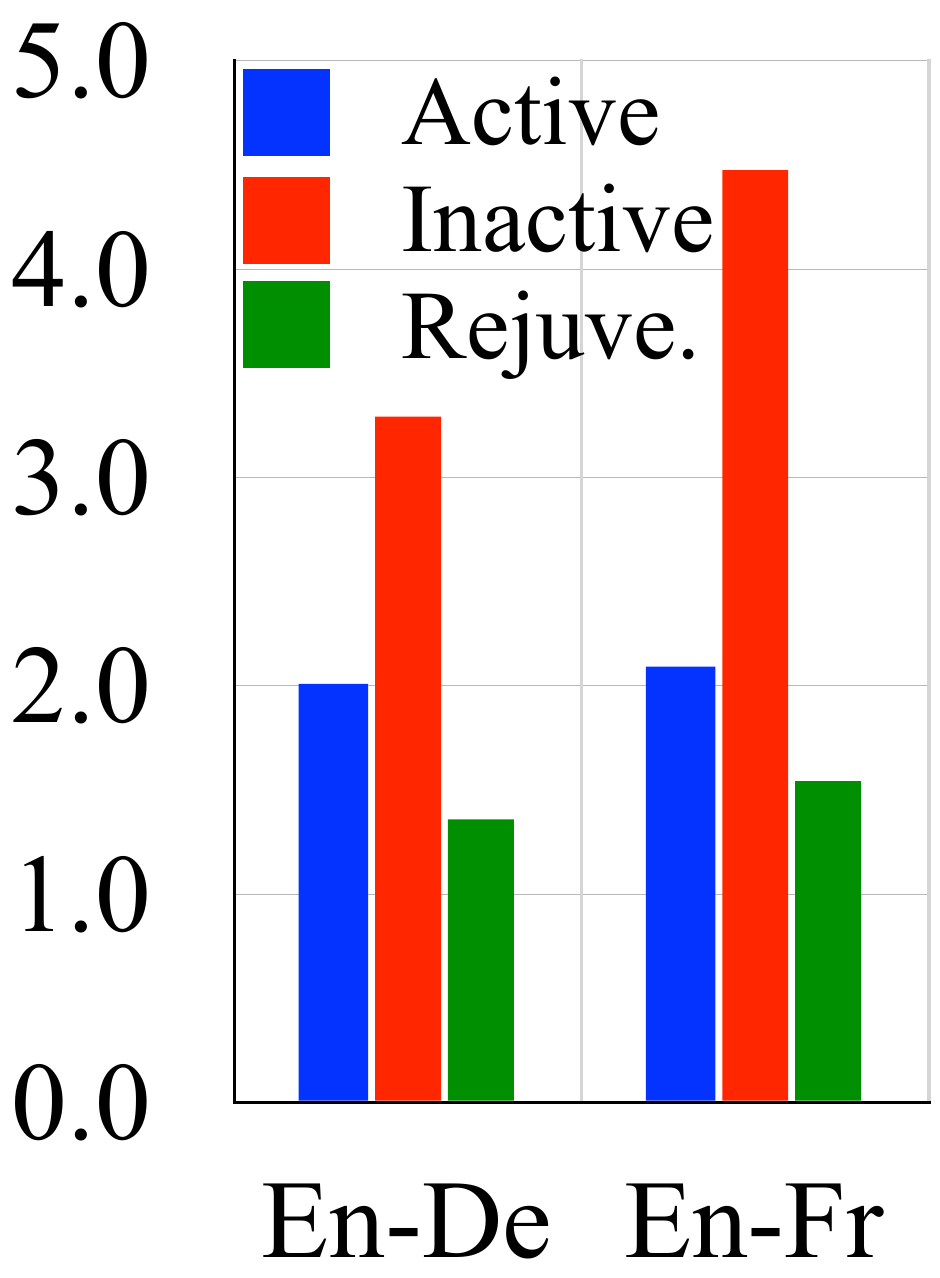}}
    \caption{Linguistic properties of different training examples: frequency rank ($\uparrow$ more difficult), coverage ($\downarrow$ more difficult), and uncertainty ($\uparrow$ more difficult).}
    \label{fig:linguistic-properties}
\end{figure}

In this section, we investigated the linguistic properties of the identified inactive examples.
We explored the following 3 types of properties: frequency rank, coverage, and uncertainty. Frequency rank measures the rarity of words, which is calculated for the target words since the proposed data rejuvenation method modifies the target side of the training examples. Coverage measures the ratio of source words being aligned by any target words. Uncertainty measures the level of multi-modality of a parallel corpus~\cite{zhou2019understanding}. 
These properties reflect the difficulty of training examples to be learned by NMT models.

Figure~\ref{fig:linguistic-properties} depicts the results. As seen, the linguistic properties consistently suggest that inactive examples are more difficult than those active ones. By rejuvenation, the inactive examples are transformed into much simpler patterns such that NMT models are able to learn from them.

\subsection{Learning Stability}
\label{sec:learning}

\begin{figure}[t]
    \centering
    \subfloat[Training Loss]{        \includegraphics[height=0.28\textwidth]{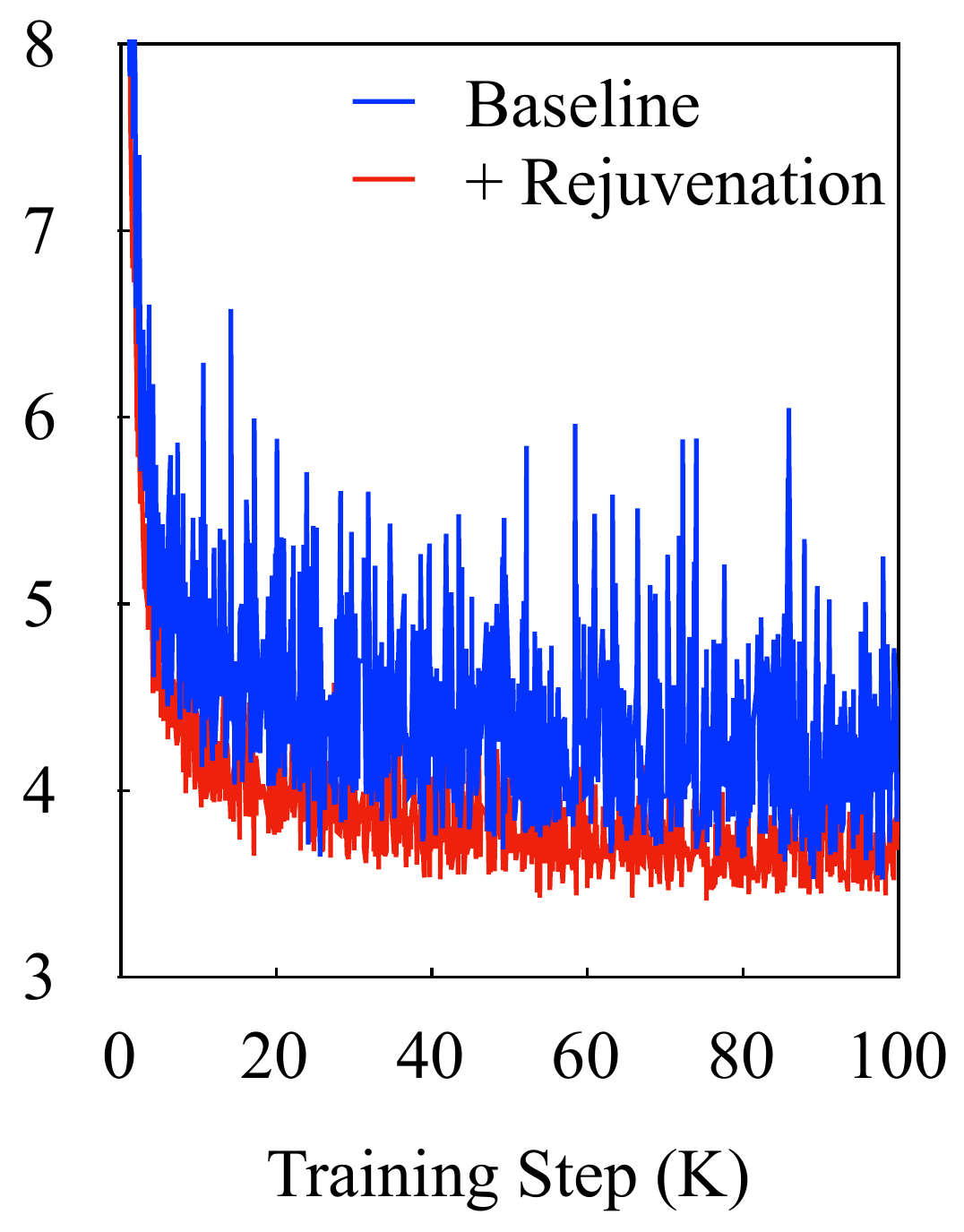}} 
    ~~~
    \subfloat[Validation BLEU]{        \includegraphics[height=0.28\textwidth]{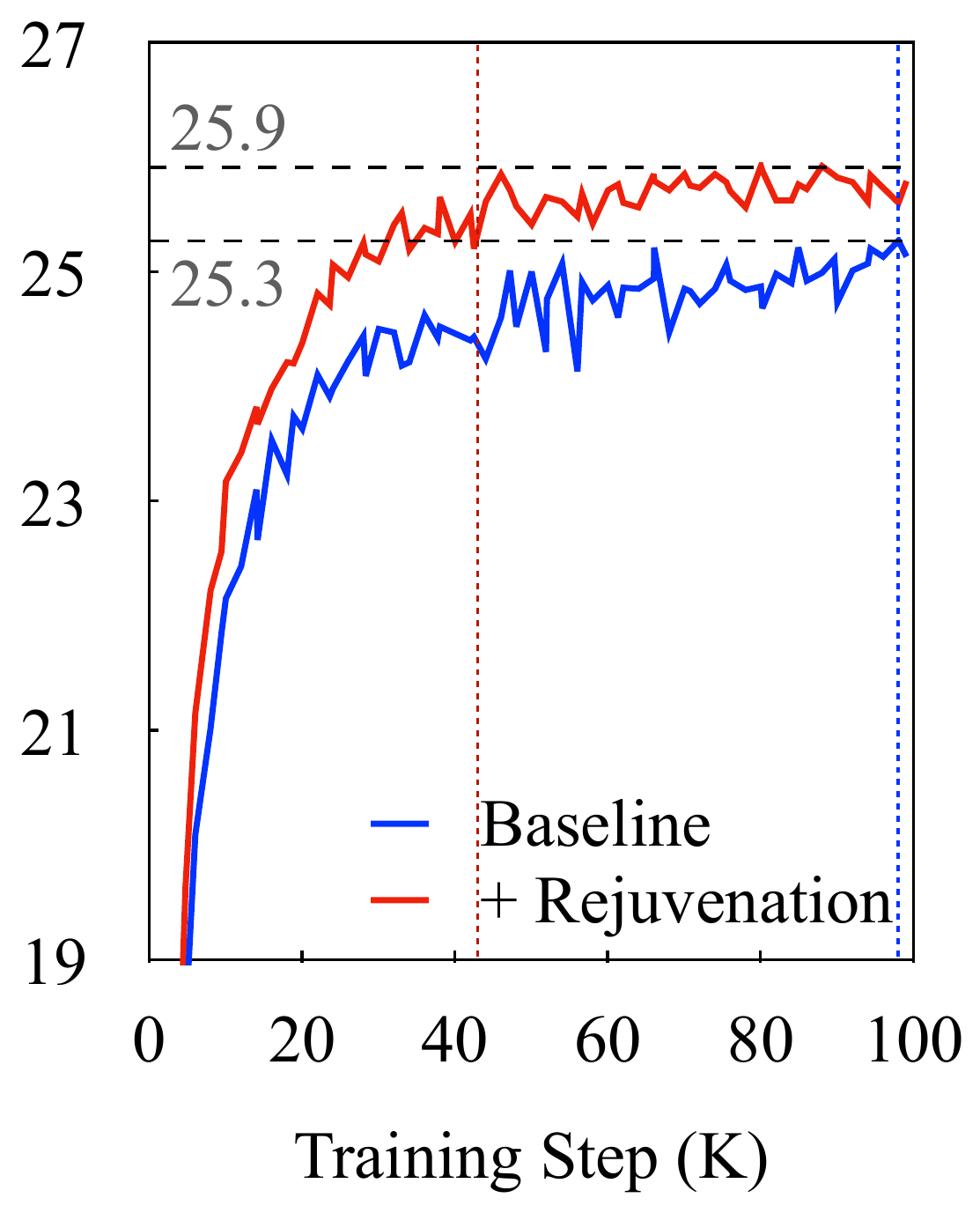}}
    \caption{Learning curves on the En$\Rightarrow$De dataset.} 
    \label{fig:analysis-trainloss-validbleu}
\end{figure}

In this section, we studied how data rejuvenation improved translation performance from the perspective of the optimization process, as shown in Figure~\ref{fig:analysis-trainloss-validbleu}.
Concerning the training loss (Figure~\ref{fig:analysis-trainloss-validbleu}(a)), our approach converges faster and presents much less fluctuation than the baseline model during the whole training process. 
Correspondingly, the BLEU score on the validation set is significantly boosted (Figure~\ref{fig:analysis-trainloss-validbleu}(b)). 
These results suggest that data rejuvenation is able to accelerate and stabilize the training process.

\subsection{Generalization Capability}
\label{sec:generalization}

\begin{table}[t]
\fontsize{10}{11}\selectfont
\centering
\begin{tabular}{l| c |  c}
    {\bf Model} & {\bf Margin} & {\bf GSNR} \\
    \hline \hline
    \textsc{Transformer-Base}  &  0.68  &  5.2e-3  \\
    \hline
    \em~~~ + Data Rejuvenation &  0.71  &  8.5e-3   \\
    \end{tabular}
    \caption{Results of generalization capability on the En$\Rightarrow$De dataset. 
    Larger Margin/GSNR values denote better generalization capability.
    }
    \label{tab:generalization}
\end{table}

In this section, we investigated how data rejuvenation affected the generalization capability of NMT models with two measures, namely, Margin~\cite{bartlett:2017:NeurIPS} and Gradient Signal-to-Noise Ratio~\citep[GSNR,][]{liu:2020:ICLR}. 
Table~\ref{tab:generalization} lists the results, in which the GSNR values are at the same order of magnitude as that reported by~\newcite{liu:2020:ICLR}. 
As seen, our approach achieves noticeably larger Margin and GSNR values, demonstrating that data rejuvenation improves the generalization capability of NMT models.

\subsection{Speeding Up}
\label{sec:speedup}

The pipeline of data rejuvenation in Figure~\ref{fig:framework} is time-consuming: 
training the identification and rejuvenation models in sequence as well as the scoring and rejuvenating procedures make the time cost of data rejuvenation more than 3X that of the standard NMT system.
To save the time cost, a promising strategy is to let the identification model take the responsibility of rejuvenation. Therefore, we used the \textsc{Transformer-Big} model with the large batch configuration trained on the raw data to accomplish both identification and rejuvenation. The resulted data is used to train two final models, i.e., \textsc{Transformer-Big} and \textsc{DynamicConv}. 

Figure~\ref{tab:speedup} lists the results. With almost no decrease of translation performance, the time cost of data rejuvenation is reduced by about 33\%. This makes the total time cost comparable with those data manipulation or augmentation techniques that require additional NMT systems, such as data diversification~\cite{nguyen2019data} and back-translation~\cite{Sennrich:BT}. In addition, the superior performance of \textsc{DynamicConv} (i.e., 30.4) further demonstrates the high agreement of inactive examples across architectures.

\begin{table}[t]
\fontsize{10}{11}\selectfont
\centering
\begin{tabular}{c | c r | c r}
    \multirow{2}{*}{\bf Method} & \multicolumn{2}{c|}{\bf \textsc{Trans.-Big}} &  \multicolumn{2}{c}{\bf \textsc{Dyn.Conv}} \\
    \cline{2-5}
        &   BLEU    &   Time    &   BLEU    &   Time\\
    \hline \hline
    Standard       &   29.6   & 32h   &   29.7    & 31h  \\
    \hline
    Rejuvenate     &   30.3   & +65h  &   30.2    & +62h  \\
    \hdashline
    Rej.--Big     &   30.2  &  +33h  &  30.4  &  +32h \\
    \end{tabular}
    \caption{Results of speeding up (``Rej.--Big'') on the WMT14 En$\Rightarrow$De dataset. ``Time'' denotes the time of the whole process using 4 NVIDIA Tesla V100 GPUs.}
    \label{tab:speedup}
\end{table}

\subsection{Analysis on Inactive Examples}

\begin{figure}[t]
    \centering
    \subfloat[Probability]{   \includegraphics[height=0.27\textwidth]{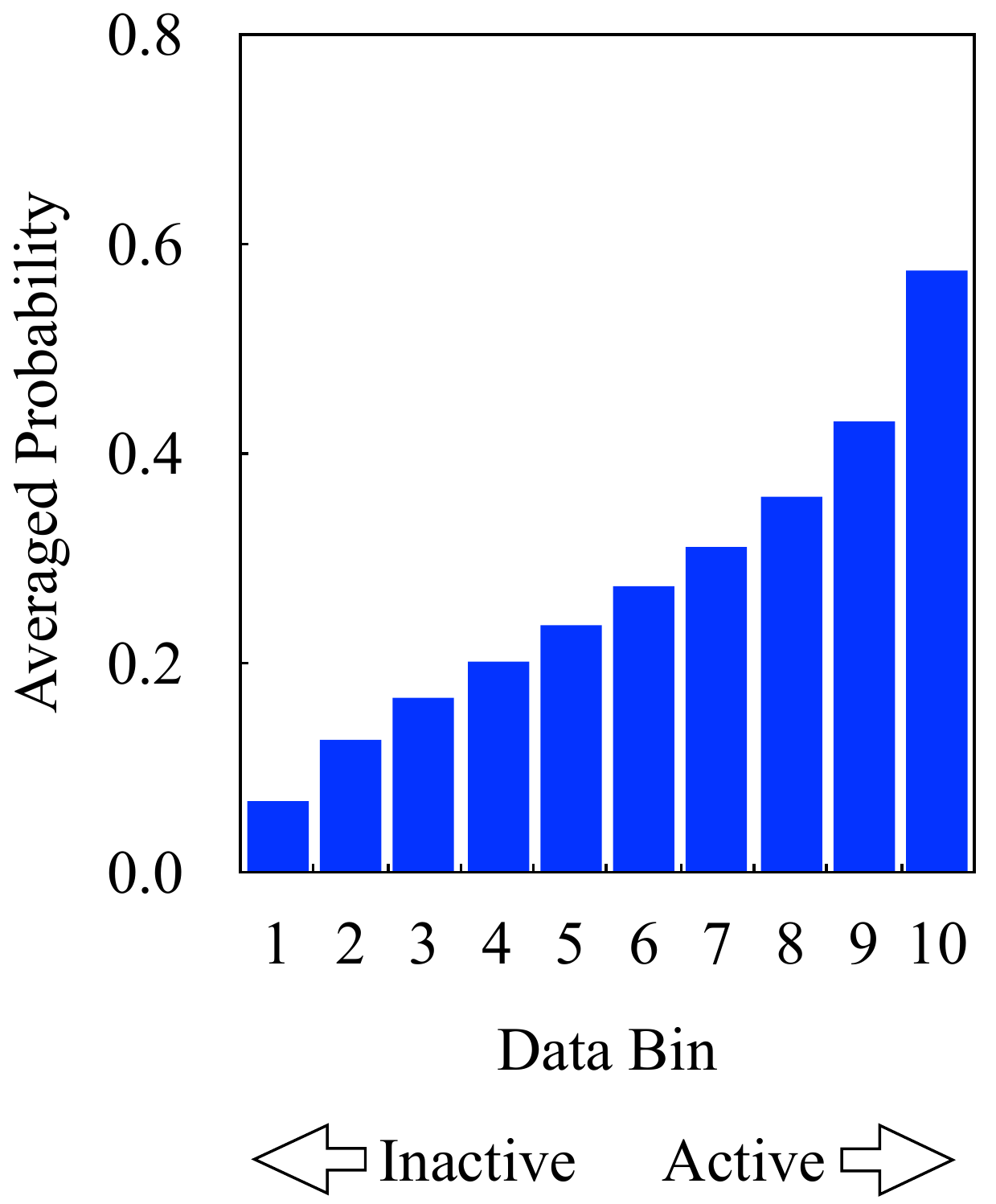}}
    \hfill
    \subfloat[Source-Translated Ratio]{   \includegraphics[height=0.27\textwidth]{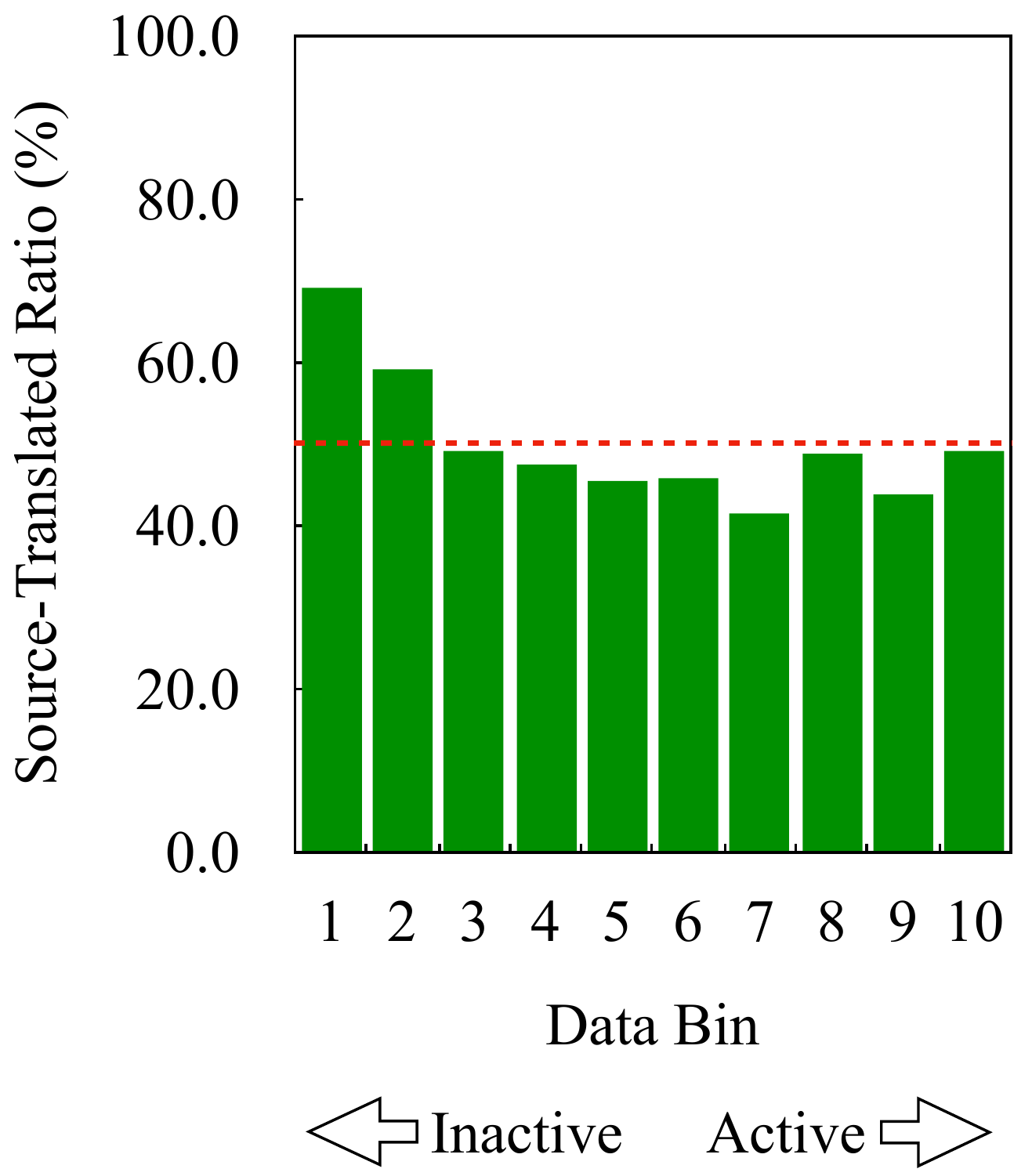}}
    \caption{Probability and ratio of source-translated examples over the data bins of En$\Rightarrow$De test set.}
    \label{fig:translationese-diagram}
\end{figure}

\paragraph{Human Translations from Target to Source as Inactive Examples?}
Since forward translation performs better than back-translation for rejuvenation, one would wonder if the inactive examples correspond to human translations from target to source. For simplicity, we name such examples as source-translated whereas source-natural otherwise. The information of source-translated/natural examples is unavailable for training examples, but fortunately is provided for test sets\footnote{\url{https://www.statmt.org/wmt14/test-full.tgz}}. 
We split the test examples of En$\Rightarrow$De into 10 data bins according to the sentence-level probability (see Eq.~(\ref{eq_inactive_metric})) of the identification model (i.e., \textsc{Transformer-Base}), and then calculate the ratio of source-translated examples in each bin. As seen in Figure~\ref{fig:translationese-diagram}, the ratios of source-translated examples in $1^{st}$ and $2^{nd}$ bins (i.e., 69\% and 59\%) significantly exceed that in the whole test set (i.e., 1500/3003), suggesting that human translations from target to source are more likely to be inactive examples.

\paragraph{Case Study.}
By inspecting the inactive examples, we find that the target sentences tend to be paraphrases of the source sentences rather than direct translations. We provide two cases in Table~\ref{tab:case-study}. In the first case, the target sentence does not translate ``finished the destruction of the first" in the source sentence directly but rephrases it as ``tat dann das seine und zerstörte den Rest", meaning ``then did his and destroyed the rest" (that was not destroyed by The First World War). As for the second case, ``denied by the latter" uses passive voice but its corresponding phrase in the target sentence is in active voice. These observations indicate that the inconsistent structure or expression between source and target sentences could make the examples difficult for NMT models to learn well.

\begin{table}[t]
\small
\centering
\resizebox{0.98\columnwidth}{!}{
\begin{threeparttable}
\begin{tabular}{p{0.2cm}|p{0.5cm}|p{6cm}}
 & \bf Side & \bf \qquad\qquad\qquad\quad~Sentence \\
\hline
\hline
\multirow{10}{*}{\rotatebox[origin=c]{90}{En$\Rightarrow$De}}
& \multirow{2}{*}{\texttt{X}} &  The Second World War \uline{finished the destruction of the first} . \\
\cline{2-3}
& \multirow{4}{*}{\texttt{Y}} & Der zweite Weltkrieg \uline{tat dann das seine und zerstörte den Rest} . \\
&  & \texttt{=>En:} The Second World War {\color{red}\uline{then did his and destroyed the rest}} . \\
\cline{2-3}
& \multirow{4}{*}{\texttt{Y'}} & Der Zweite Weltkrieg \uline{beendete die Zerstörung des ersten} . \\
&  & \texttt{=>En:} The Second World War {\color{blue}\uline{ended the destruction of the first}} . \\
\hline
\hline
\multirow{10}{*}{\rotatebox[origin=c]{90}{En$\Rightarrow$Fr}}
& \multirow{2}{*}{\texttt{X}} &  Anything \uline{denied by the latter} was effectively confirmed as true . \\
\cline{2-3}
& \multirow{4}{*}{\texttt{Y}} & Tout ce \uline{que démentait cette agence} se révélait dans la pratique bien réel . \\
&  & \texttt{=>En:} Everything that {\color{red}\uline{this agency denied}} turned out to be very real in practice . \\
\cline{2-3}
& \multirow{4}{*}{\texttt{Y'}} & Toute chose \uline{niée par ce dernier} a été effectivement confirmée comme vraie . \\
&  & \texttt{=>En:} Anything {\color{blue}\uline{denied by the latter}} has actually been confirmed to be true . \\
\end{tabular}
\end{threeparttable}
}
\caption{Inactive examples from the training sets of En$\Rightarrow$De and En$\Rightarrow$Fr. \texttt{X}, \texttt{Y} and \texttt{Y'} represent the source sentence, target sentence, and the rejuvenated target sentence, respectively. \texttt{Y} and \texttt{Y'} are also translated into English (\texttt{=>En:}) by Google Translate for reference. For either example, the underlined phrases correspond to the same content.}
\label{tab:case-study}
\end{table}

\section{Conclusion}

In this study, we propose data rejuvenation to exploit the inactive training examples for neural machine translation on large-scale datasets. The proposed data rejuvenation scheme is a general framework where one can freely define, for instance, the identification and rejuvenation models. Experimental results on different model architectures and language pairs demonstrate the effectiveness and universality of the data rejuvenation approach.

Future directions include exploring advanced identification and rejuvenation models that can better reflect the learning abilities of NMT models, as well as validating on other NLP tasks such as dialogue and summarization.

\section*{Acknowledgments}

This work is partially supported by the National Key Research and Development Program of China (No. 2018AAA0100204) and the Research Grants Council of the Hong Kong Special Administrative Region, China (No.~CUHK 14210717 of the General Research Fund). We sincerely thank Yongchang Hao for the help in implementing \textsc{Lstm} in the \textsc{Transformer} framework, Shuo Wang for the valuable advice on technical details, and the anonymous reviewers for their insightful suggestions on various aspects of this work.

\bibliography{emnlp2020}
\bibliographystyle{acl_natbib}

\appendix
\section{Appendix}
\label{sec:appendix}

\subsection{Model Implementation}
We adopt the default implementation of models in Fairseq\footnote{\url{https://github.com/pytorch/fairseq}}~\cite{ott:2019:naacl} except for \textsc{Lstm}. 

\paragraph{\textsc{Lstm}.}
We follow~\newcite{domhan2018much} to implement \textsc{Lstm} by replacing the self-attention (SAN) layers in \textsc{Transformer-Base} with \textsc{Lstm} layers. Specifically, we use a bidirectional \textsc{Lstm} for each layer of the encoder, and a unidirectional \textsc{Lstm} for each layer of decoder. Each bidirectional \textsc{Lstm} layer is followed by a fully-connected layer with \texttt{ReLU} as the activation function.

Note that the training strategies of models with the proposed \emph{data rejuvenation} are the same as that of the corresponding baseline models, without any modification of hyper-parameters.

\subsection{Linguistics Properties}
To understand the characteristics of inactive examples, we compare them with active examples and rejuvenated examples in terms of 3 linguistics properties: frequency rank, coverage, and uncertainty.

\paragraph{Frequency Rank.}
Frequency rank measures the rarity of words, which is calculated for the target words since our proposed data rejuvenation method modifies the target side of the training examples. In the target vocabulary, words are sorted in the descending order of their frequencies in the whole training data, and the frequency rank of a word is its position in the dictionary. Therefore, the higher the frequency rank is, the more rare the word is in the training data. We report the averaged frequency rank of each of the three subsets. The larger frequency rank of inactive examples indicates that they contain more rare words, which make them more difficult to be learned by NMT models than the active examples.  

\paragraph{Coverage.}
Coverage measures the ratio of source words being aligned by any target words~\cite{tu2016modeling}. Firstly, we train an alignment model on the training data by \emph{fast-align}\footnote{\url{https://github.com/clab/fast\_align}}~\cite{dyer2013simple}, and force-align the source and target sentences of each subset. Then, we calculate the coverage of each source sentence, and report the averaged coverage of each subset. The lower coverage of inactive examples indicates that they are not very well aligned as the active examples, which also make them more difficult for NMT models to learn.

\paragraph{Uncertainty.}
Uncertainty measures the level of multi-modality of a parallel corpus~\cite{zhou2019understanding}. The uncertainty of a source sentence can reflect the number of its possible translations in the target side. We consider the corpus level uncertainty, which measures the complexity of each subset. Corpus level uncertainty is simplified as the sum of entropy of target words conditioned on the aligned source words denoted $H(y|x = x_t)$. Therefore, an alignment model is also required. To prevent uncertainty from being dominated by frequent words, we follow \newcite{zhou2019understanding} to calculate uncertainty by averaging the entropy of target words conditioned on a source word denoted $\frac{1}{|\mathcal{V}_x|}\sum_{x\in\mathcal{V}_x}H(y|x)$. The larger uncertainty of inactive examples indicates that there are more possible translations for each source sentence of them. That is to say, inactive examples contain more complex patterns, which are more difficult to be learned by NMT models.

\begin{table*}[t]
\centering
\begin{tabular}{c|l||l| l c||l| l c}
\multirow{2}{*}{\bf System}  &   \multirow{2}{*}{\bf Architecture}  & \multicolumn{3}{c}{\bf WMT14 En$\Rightarrow$De}  &  \multicolumn{3}{||c}{\bf WMT14 En$\Rightarrow$Fr}\\
    \cline{3-8}
        &   &  \em valid  &  \em test  & $\bigtriangleup$    &  \em valid  & \em test & $\bigtriangleup$ \\
    \hline \hline
    \multicolumn{8}{c}{{\em Our NMT Systems}}   \\ 
    \hline
    \multirow{10}{*}{\em This work}
    &   \textsc{Lstm}              &  25.3  &    26.5   &   --  &  33.4  &   40.6  &   --\\
    &   ~~~ +  Data Rejuvenation   &  26.1  &    27.0$^\uparrow$       &    +0.5   &  33.8  &  41.1$^\uparrow$  &   +0.5\\ 
    \cline{2-8}
    &   \textsc{Transformer-Base}  &  26.3  &  27.5  & -- &  33.0  &  40.2  &   -- \\
    &   ~~~ +  Data Rejuvenation   &  26.8  &  28.3$^\Uparrow$  & +0.8  &  33.2  &  41.0$^\Uparrow$ & +0.8 \\ 
    \cdashline{2-8}
    &   \textsc{Transformer-Big}  &  26.9  &  28.4  & -- &  34.5  &   42.4 &   -- \\
    &   ~~~ +  Data Rejuvenation  &  27.3  &  29.2$^\Uparrow$  &  +0.8   &  34.9  & 43.0$^\uparrow$  &  +0.6 \\ 
    \cdashline{2-2} \cdashline{3-5} \cdashline{6-8}
    &   ~~~ + Large Batch  & 27.4 &   29.6     & -- &  35.0  &  43.5  &   -- \\
    &   ~~~~~~~~ +  Data Rejuvenation  &  28.0  &  30.3$^\Uparrow$  &  +0.7 &  35.4  & 44.0$^\uparrow$ & +0.5  \\
    \cline{2-8}
    &   \textsc{DynamicConv}      & 27.2 &   29.7             & --     &  35.0  & 43.3 &   -- \\
    &   ~~~ +  Data Rejuvenation  & 27.6 &   30.2$^\uparrow$  &  +0.5  &  35.2  & 43.9$^\uparrow$  &  +0.6 \\
  \end{tabular}
  \caption{Translation performance of \emph{valid} and \emph{test} sets across model architectures and language pairs.} 
  \label{tab:main_details}
\end{table*}

\subsection{Generalization Capability}

\paragraph{Margin.}
Margin~\cite{bartlett:2017:NeurIPS} is a classic concept in support vector machine, measuring the geometric distance between the support vectors and the decision boundary. To apply margin for NMT models, we follow \newcite{li2019understanding} to compute word-wise margin, which is defined as the probability of the correctly predicted word minus the maximum probability of other word types. We compute the word-wise margin over the training set and report the averaged value. 

\paragraph{GSNR.}
The gradient signal to noise ratio (GSNR) metric~\cite{liu:2020:ICLR} is proposed to positively correlate with generalization performance.
The calculation of a parameter's GSNR is defined as the ratio between its gradient's squared mean and variance over the data distribution. For NMT models, we compute GSNR of each parameter and report the averaged value over all the parameters.

Compared with the baseline model trained on the raw data, the model trained with our \emph{data rejuvenation} has larger Margin and GSNR, suggesting that \emph{data rejuvenation} is able to improve the generalization capability of the final NMT models.

\subsection{Validation Performance}

In Table~\ref{tab:main_details}, we provide details of the main results, including the translation performance on both the validation and test sets. 
Generally, the models with our \emph{data rejuvenation} outperform the baseline models on both validation and test sets.

\subsection{More Ablation Studies}

\paragraph{Reversed Models for Identification and Rejuvenation.}
Some researchers are curious whether the back-translation strategy will work if reversed NMT models are adopted for both identification and rejuvenation. To study this strategy, we trained a reversed translation model on the raw data as the identification model, and another reversed translation model on the identified active examples as the rejuvenation model. Finally, we trained a forward translation model on the rejuvenated training data. 
The final model marginally outperforms the baseline (27.7 v.s. 27.5) but significantly underperforms the forward translation method (27.7 v.s. 28.3).

\paragraph{Fine-tuning on Inactive Examples.}
We also tried a more straightforward strategy to re-use the inactive examples, i.e., to fine-tune the baseline NMT models on the inactive examples. We investigated this strategy on the En$\Rightarrow$De dataset with a pre-trained \textsc{Transformer-Base} model. Experimental results show that the model diverges after fine-tuning on the inactive examples either individually or in combination with similar-sized active examples (the latter diverges slower), suggesting that fine-tuning on the inactive examples may not be a promising strategy.

\subsection{Doubts on Main Results}

\paragraph{Random Seeds.}
Some researchers may doubt if the improvement achieved by our approach comes from lucky random starts.
To dispel this doubt, we conducted experiments on the En$\Rightarrow$De dataset using the \textsc{Transformer-Base} model with three random seeds (i.e., 1, 12, and 123). Our approach consistently outperforms the baseline model in all cases (i.e., 27.5/28.3, 27.4/28.2, and 27.1/27.9), demonstrating the effectiveness of our approach.

\paragraph{Source Language.}
Some researchers may have questions about the language pairs used in the experiments that both language pairs have English as the source language, which could determine the rejuvenation strategy.
To demonstrate the universality of our approach across language directions, we conducted an experiment on the WMT14 De-En translation task. The \textsc{Transformer-Base} model achieved a BLEU score of 31.2, and the data rejuvenation approach improves performance by +0.6 BLEU point.

\end{document}